\documentclass[aps,preprint]{revtex4}
\usepackage{amsmath,amssymb,amsthm,mathrsfs,amsfonts,dsfont} 
\usepackage{graphicx}
\usepackage{subfigure}
\usepackage{amsbsy}
\usepackage{bm}

\def\t{\mathrm{T}}

\def\id{\mathds{1}}

\begin{document}

\title{Fermi-Bose Machine achieves both generalization and adversarial robustness}
\author{Mingshan Xie$^{1}$}
\author{Yuchen Wang$^{1}$}
\author{Haiping Huang$^{1,2}$}
\email{huanghp7@mail.sysu.edu.cn}
\affiliation{$^{1}$PMI Lab, School of Physics,
Sun Yat-sen University, Guangzhou 510275, People's Republic of China}
\affiliation{$^{2}$Guangdong Provincial Key Laboratory of Magnetoelectric Physics and Devices,
Sun Yat-sen University, Guangzhou 510275, People's Republic of China}
\date{\today}

\begin{abstract}
	Distinct from human cognitive processing, deep neural networks trained by backpropagation can be easily fooled by adversarial examples. To design a semantically meaningful representation learning, we discard backpropagation, and instead, propose a local contrastive learning, where the representations for the inputs bearing the same label shrink (akin to boson) in hidden layers, while those of different labels repel (akin to fermion). This layer-wise learning is local in nature, being biological plausible. A statistical mechanics analysis shows that the target fermion-pair-distance is a key parameter. Moreover, the application of this local contrastive learning to MNIST benchmark dataset demonstrates that the adversarial vulnerability of standard perceptron can be greatly mitigated by tuning the target distance, i.e., controlling the geometric separation of prototype manifolds. 
\end{abstract}

 \maketitle

\section{Introduction}
Deep neural networks (DNNs) are widely used in current artificial intelligence studies, including Chat GPT and generative diffusion models, producing remarkable revolution in many fields of science and technology~\cite{DL-2016,Sparks-2023,Sora-2024}. These networks are commonly trained by backpropagation algorithms, which are criticized to be far from biological intelligence~\cite{BPbrain-2020}, mainly because a global cost-function should be defined and each weight update requires propagating errors. In addition, a severe caveat of DNNs is their adversarial vulnerability~\cite{Szegedy-2014,Good-2015}, i.e., a human-imperceptible perturbation to an input can completely change the classification of that input. There do not yet exist effective ways to control this vulnerability. One standard strategy in practice is the adversarial training which takes adversarial examples as a part of training data~\cite{Madry-2018}. But a trade-off between standard accuracy and robustness might be inevitable~\cite{adv-2018d,Precise-2022}.

We resolve these two challenges by a unified principle formulated as a statistical mechanics problem. Semantically meaningful representations must emerge during a good representation learning~\cite{Huang-2023}, as commonly observed in information disentangling process in the ventral stream of visual cortex~\cite{Dicarlo-2007,DiCarlo-2012}. We argue that this semantic hierarchy of concepts can by learned by designing a contrastive Hamiltonian. The data pair belonging to the same class forms a boson pair, as expected from physics that the Euclidean distance between the pair must be minimized, while the pair belonging to different classes forms a fermion pair, and the distance must be enlarged (a physics intuition). Both types of distance construct the Hamiltonian, which can be trained layer by layer and thus backpropagating a global error is not needed. This principle, namely Fermi-Bose machine (FBM), realizes geometry separation of internal representations in the latent space of DNNs, bearing similarity with recent empirical observation of hierarchical nucleation in DNNs~\cite{Laio-2020}, latent space clustering~\cite{SSL-2018} and emergence of compact latent space~\cite{Jiang-2021c,Li-2023}.

Remarkably, \textit{without} adversarial training,  our FBM demonstrates the ability to mitigate the adversarial vulnerability by tuning only a target fermion-pair distance, aligning with a recent hypothesis of relationship between data concentration and adversarial robust classifier~\cite{Adv-2023}. Our work shows that the geometrically separated representations encode semantic hierarchy that facilitates \textit{both} discrimination and robustness against class-preserving perturbations, and moreover, this can be realized by a local learning, amenable to a statistical mechanics analysis.

\section{Machine setting}
We consider a classification task implemented by a deep network with $L$ layers. $N_\ell$ ($\ell=1,\ldots,L$) denotes the layer size. $L-2$ hidden layers are individually learned by our FBM. We first consider a single layer perceptron of $N$-$1$ structure ($N_1=N$ and $N_2=1$) for statistical mechanical analysis, which can be straightforward to extend to more layers and multiple labels. In the toy theoretical model, the data of Gaussian mixture are considered. We then test FBM in a three layer network for classifying real benchmark dataset. In the toy model, we analyze only the representation learning, while in learning real dataset, we add a final readout layer to leverage well-separated representations for classification. 

The network parameters are specified by all weight matrices between layers. The weight value from neuron $j$ at the upstream layer $\ell-1$ to neuron $i$ at the downstream layer $\ell$ is denoted by $w_{ij}^\ell$. The activation of neuron $i$ in the layer $\ell$ is given by $h^\ell_i = \phi\left(z^\ell_i\right)$, where $z^{\ell}_i = \sum^{N_{\ell-1}}_{j=1} w^{\ell}_{ij}h^{\ell-1}_j$, and $\phi(\cdot)$ is the nonlinear function $\mathrm{tanh}(\cdot)$. In the toy model setting, the hidden layer has a single neuron, and thus the subscript of $z$ in the following analysis indicates which input of the fermion (or boson) pair.

\begin{figure}
    \centering
\includegraphics[width=0.8\textwidth]{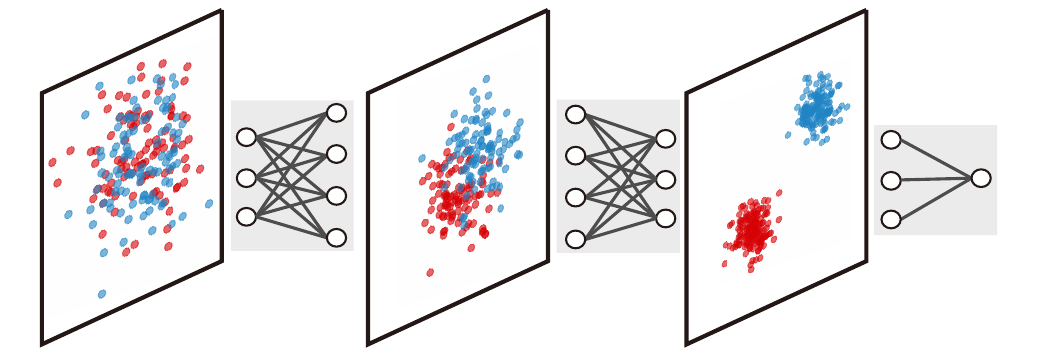}
        \caption{Schematic illustration of FBM. A network with structure 3-4-3-1 is used to learn geometrically separated representations, which become increasingly compact per class, such that the final linear separation is achievable. A layer-wise training is carried out, and thus the learning is local.  } 
        \label{fig1}
    \end{figure}

To learn a meaningful representation at each layer, a layer-wise training is carried out. The inputs are decomposed into boson pair and fermion pair (this is only a physics metaphor but not talking about real physical particles). This process is akin to data augmentation. The dataset is defined by  $\mathcal{D}\equiv\{\left(\mathbf{x}^{\mu}_1, \mathbf{x}^{\mu}_2, \sigma^{\mu}\right)\}_{\mu\in[P]}$, where  $\mathbf{x}_1^\mu,\mathbf{x}_2^\mu \in \mathbb{R}^N$ and $\sigma^\mu \in \{0, 1\}$. $\sigma^\mu = 1$ if $\mathbf{x}^{\mu}_1, \mathbf{x}^{\mu}_2$ belong to the same category (boson pair), while $\sigma^\mu = 0$ if $\mathbf{x}^{\mu}_1, \mathbf{x}^{\mu}_2$ belong to different categories (fermion pair). The loss function can thus be constructed below:
    \begin{equation}\label{Hami}
    \begin{aligned}
      \mathcal{L} =\sum_{\mu=1}^P\frac{1}{2} \left[\sigma^\mu D_\mu^2+(1-\sigma^\mu) \varphi\left(d_F - D_\mu ^2\right) \right]+\frac{\lambda_w}{2}\Vert\mathbf{w}\Vert^2_2,
      \end{aligned}
    \end{equation}
    where $\lambda_w$ is a weight regularization parameter, $\sigma^\mu=\frac{1+y_1^\mu y_2^\mu}{2}$, $y_1^\mu$ and $y_2^\mu$ are the labels of $\mathbf{x}^\mu_1$ and $\mathbf{x}^\mu_2$, respectively, and $D_\mu^2 = [\phi(z_1^\mu)-\phi(z_2^\mu)]^2$ is the Euclidean distance. $\varphi(\cdot)$ denotes a relu-like function. The first term encourages contraction of boson pairs,  while the second term repels fermion pairs, and $d_F$ sets the target distance of fermion pairs. An illustration of this idea is given in Fig.~\ref{fig1}. One can also add the activity regularization into the above loss function for more biological sparse activation. The layer-wise Hamiltonian in Eq.~\eqref{Hami} can also be adapted to unsupervised learning where positive pairs (the transformation of one input and the input itself) and negative pairs (two different inputs from the training batch) are used~\cite{SCL-2020}.
    
For the sake of statistical mechanics analysis, we use the Gaussian mixture data, which is defined by $\mathrm{P}\left(\mathbf{x}_1,\mathbf{x}_2, y_1,y_2\right) = \mathrm{P}\left(\mathbf{x}_1\vert y_1\right) \mathrm{P}\left(\mathbf{x}_2\vert y_2\right)\mathrm{P}\left(y_1,y_2\right)$,  where $\mathrm{P}\left(y_1,y_2\right) = \frac{1}{2}\rho\delta(y_1-y_2) +  \frac{1}{2}\left(1 - \rho\right)\delta(y_1+y_2) $. $\rho$ specifies the fraction of boson pairs. $\mathrm{P}\left(\mathbf{x}\vert y\right) = \mathcal{N}\left(\frac{my}{N}\id, \frac{\Delta^2}{N}\mathbf{I}\right)$, where $\mathds{1}$ is an all-one vector, $\mathbf{I}$ is an identity matrix, and $y \in \{\pm 1\}$ for two classes. $m$ and $\Delta$ determine the mean and variance of each Gaussian distribution. 

\section{Results}
\subsection{Order parameters determine the generalization performance}
We focus on the limit of large number of training pairs and input dimension, while keeping the ratio $\alpha=\frac{P}{N}$ fixed. The partition function of the model reads,
\begin{equation}\label{pf}
Z(\beta,\mathcal{D},\lambda_w,d_F)=\int d\mathbf{w} e^{-\beta\mathcal{L}(\mathcal{D},\mathbf{w},\lambda_w,d_F)},
\end{equation}
where $\beta$ denotes the inverse temperature, the partition function is data-dependent, and thus the free energy must be averaged over the data ($\langle\cdot\rangle$ below), which is highly non-trivial and we have to use the replica method~\cite{Mezard-1987,HH-2022}:
\begin{equation}
-\beta f=\langle\ln Z\rangle=\lim_{n\to0,N\to\infty}\frac{\ln\langle Z^n\rangle}{nN},
\end{equation}
where $f$ is the free energy density, and we have assumed the limit of zero replica number and thermodynamic limit are exchangeable~\cite{Mezard-1987}.

 As detailed in Methods, the free energy density depends on physically relevant order parameters  $\{M,q,Q\}$ together with the conjugate counterparts $\{\hat{M},\hat{q},\hat{Q}\}$ as follows:
\begin{equation}\label{fRS}
\begin{aligned}
-\beta f&= \mathop{\mathrm{Extr}}\limits_{M,q, Q,\hat{M},\hat{q},\hat{Q}} {\hat{M}}{M} +{\hat{{q}}}{q} -
			\frac{{\hat{{Q}}}{Q}}{2} \\
   &+\Psi_{S}\left({\hat{Q}}, {\hat{q}}, {\hat{M}}\right) + \Psi_{E} \left({Q}, {q}, {M} \right),
\end{aligned}
\end{equation}
where the entropy term $\Psi_S\left({\hat{Q}}, {\hat{q}}, {\hat{M}}\right)=\frac{1}{2}\ln\frac{2\pi}{2\hat{q}+2\beta\lambda_w-\hat{Q}}+\frac{\hat{M}^2-\hat{Q}}{2(2\hat{q}+2\beta\lambda_w-\hat{Q})}$, and the energy term
$\Psi_{E} \left({Q}, {q}, {M} \right)=\alpha\mathbb{E}_\mathbf{y}\int D\mathbf{v}\ln\int D\mathbf{u}e^{-\beta\mathcal{H}\left(y_1,y_2,z_1(y_1,u_1,v_1),z_2(y_2,u_2,v_2)\right)}$, where $\mathbf{y}=(y_1,y_2)$, $\mathbf{v}=(v_1,v_2)$, $\mathbf{u}=(u_1,u_2)$, $Dv$ is a Gaussian measure, and the effective Hamiltonian for the FBM reads $\mathcal{H}=\frac{1}{2}\left[\frac{1+y_1y_2}{2}D^2+\frac{1-y_1y_2}{2}\varphi(d_F-D^2)\right]$, where $D^2=[\phi(z_1)-\phi(z_2)]^2$, and the pre-activation $z_1$ and $z_2$ can be reparametrized by the order parameters (see Methods). To get Eq.~\eqref{fRS}, the replica symmetry (RS) ansatz is used~\cite{Mezard-1987,HH-2022}, i.e., permutation of replica index does not affect the result. $M$ characterizes the mean of learned weight values;
$q$ denotes the variance of weight values; $Q$ denotes the overlap between two typical weight values sampled from different replicas (equivalent to states in physics~\cite{Mezard-1987}). This minimal set of order parameters obeys the following iterative equations:
\begin{equation}\label{SDE-main}
	\begin{aligned}
		&\hat{q} =\frac{\alpha\beta\Delta^2}{2} \mathop{\mathbb{E}}\limits_{\mathbf{y},\mathbf{v}} \langle\langle \mathcal{H}_{11} + \mathcal{H}_{22}- \beta(\mathcal{H}_{1}^2+\mathcal{H}_{2}^2)\rangle\rangle,\\
		&\hat{Q} =  -\alpha\beta^2\Delta^2 \mathop{\mathbb{E}}\limits_{\mathbf{y},\mathbf{v}} \left[\langle\langle\mathcal{H}_{1} \rangle\rangle^2 + \langle\langle\mathcal{H}_{2} \rangle\rangle^2 \right],\\
		&\hat{M} = \alpha\beta m \mathop{\mathbb{E}}\limits_{\mathbf{y},\mathbf{v}}\left[ y_1\langle\langle \mathcal{H}_1\rangle\rangle + y_2\langle\langle\mathcal{H}_2 \rangle\rangle\right],\\
		& q = \left(2{\hat{q}}-{\hat{Q}+ \beta\lambda_w}\right)^{-1} + \frac{\hat{{M}}^2 - \hat{{Q}}}{\left({2\hat{q}}-{\hat{Q}+ \beta\lambda_w}\right)^2 },\\
		& Q = \frac{\hat{{M}}^2 - \hat{{Q}}}{\left({2\hat{q}}-{\hat{Q}+ \beta\lambda_w}\right)^2 },\\
		&{M}  = -\frac{\hat{{M}}}{\left({2\hat{q}}-{\hat{Q}+ \beta\lambda_w}\right) },
	\end{aligned}
\end{equation}
where  $\langle\langle\bullet\rangle\rangle=\frac{\int D\mathbf{u} e^{-\beta\mathcal{H}}\bullet}{\int D\mathbf{u} e^{-\beta\mathcal{H}}}$ denotes the thermal average under the effective Hamiltonian, $\mathcal{H}_{i}=\frac{\partial\mathcal{H}}{\partial z_i}$, and $\mathcal{H}_{11}$ and $\mathcal{H}_{22}$ are corresponding second derivatives. Taking the zero temperature limit ($\beta\to\infty$) makes the Boltzmann measure [i.e., $e^{\beta\mathcal{L}}$ in Eq.~\eqref{pf}] concentrate on the ground state of the energy. Technical details are given in Methods.

To support the replica symmetry ansatz, we also derive belief propagation equation to infer the weights by using cavity method~\cite{Mezard-1987,HH-2022}. 
We first define the cavity probability $P_{i\to\mu}(w_i)$ in the absence of one data pair $\mu$, while $\hat{P}_{\nu\to i}(w_i)$ denotes the probability contribution from neighboring weights of pair $\nu$ except $i$.  These two probabilities obey the following self-consistent equations under the cavity approximation in disordered system theory~\cite{Mezard-1987,HH-2022}:
\begin{equation}\label{BP}
	\begin{aligned}
		&P_{i\to\mu}\left(w_i\right)\propto e^{-\frac{\beta\lambda_w w_i^2}{2}}\prod_{\nu\in\partial i\backslash\mu}\hat{P}_{\nu\to i}\left(w_i\right),\\
		&\hat{P}_{\nu\to i}\left(w_i\right)=\int\prod_{j\in\partial\nu\backslash i}\left[\mathrm{d}w_jP_{j\to\nu}\left(w_j\right)\right]P_{\nu}\left(\mathbf{y}^\nu,\mathbf{z}^\nu\right),
	\end{aligned}
\end{equation}
where $\partial i$ ($\partial\mu$) indicates neighbors of node $i$ ($\mu$), $\mathbf{z}^\nu=(z^\nu_1, z^\nu_2)$ are the pre-activations triggered by $(\mathbf{x}_1^\nu, \mathbf{x}_2^\nu)$, and $P_{\nu}= e^{-\beta \mathcal{L}_\nu}$ ($\mathcal{L}_\nu$ is the $\nu$-th pair related loss). Making a Gaussian assumption for $P_{i\to\mu}\sim\mathcal{N}(m_{i\to\mu},v_{i\to\mu})$ and taking a large $N$ limit (see also the previous work~\cite{Aubin-2019} for committee machine), we get a simplified belief propagation (BP) that is used in our experiments. 
\begin{equation}\label{BP2}
	\begin{aligned}
		&\boldsymbol{\omega}_{\mu \rightarrow i} = \sum_{j\in\partial\mu\backslash i} m_{j\rightarrow\mu}\mathbf{x}^\mu_{j},\\
		&\mathbf{V}_{\mu\rightarrow i} =  \sum_{j\in\partial\mu\backslash i} v_{j\rightarrow\mu}\mathbf{x}^\mu_{j}(\mathbf{x}^\mu_{j})^\t,\\	
             &B_{\nu \to i} = (\mathbf{x}_i^\nu)^\t\mathbf{f}_\nu,\\
		&A_{\nu\to i}=-(\mathbf{x}_i^\nu)^\t\frac{\partial\mathbf{f}_\nu}{\partial\boldsymbol{\omega}_{\nu\to i}}\mathbf{x}_i^\nu,\\
		&m_{i \to \mu}= \frac{\sum_{\nu\neq\mu}B_{\nu\to i}}{\beta\lambda_w+\sum_{\nu\neq\mu}A_{\nu\to i}}, \\	
		&v_{i \to \mu} =\frac{1}{\beta\lambda_w+\sum_{\nu\neq\mu}A_{\nu\to i}},
		\end{aligned}
		\end{equation}
where $\mathbf{x}^\mu_i = \left[{x}^{\mu}_{1,i}, {x}^{\mu}_{2,i}\right]^\t$, $\boldsymbol{\omega}$ is similarly defined, and a modified measure $\tilde{P}_\mu=\frac{1}{\tilde{Z}_\mu}e^{-\frac{1}{2}(\mathbf{z}^\mu-\boldsymbol{\omega}_{\mu\to i})^\t V^{-1}_{\mu\to i}(\mathbf{z}^\mu-\boldsymbol{\omega}_{\mu\to i})}P_\mu$, from which we derive $\mathbf{f}_{\mu}\equiv\frac{\partial\ln\tilde{Z}_\mu}{\partial\boldsymbol{\omega}_{\mu\to i}}$. Detailed derivations are given in Methods.

\begin{figure}
	\centering
	\includegraphics[width=0.8\textwidth]{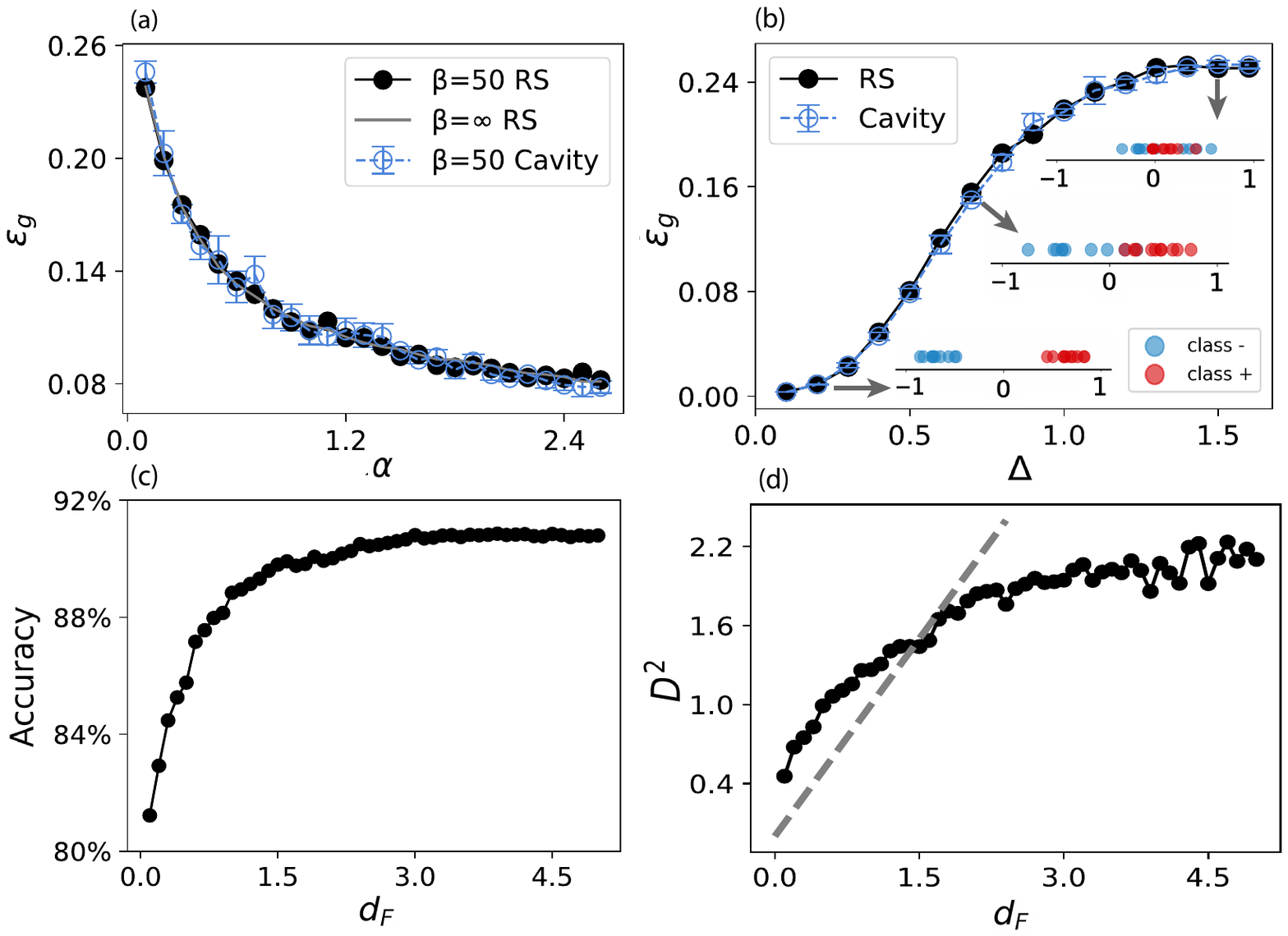}
	\caption{Generalization loss $\epsilon_g$ of FBM as a function of $\alpha$ (a) and $\Delta$ (b).  $m = 1, \rho = 0.5, d_F=1$, and $\lambda_w=0.05$. Theory matches well experiments ($N=200$).
	Error bars are computed from five independent runs. $ \Delta=0.5$ in (a). $\alpha=2.5$ and $\beta=50$ in (b). The insets from bottom to top describe the distribution of sensory inputs in the output representation space learned by BP with $\Delta=0.2, 0.7$, and $1.5$, respectively. Blue points indicate the class $y=-1$, and red points indicate $y=1$. (c) Test accuracy of FBM vs $d_F$. (d) Fermion pair distance vs $d_F$ (dashed line indicates equality). $\alpha=2.5, \Delta=0.5$ in (c,d).
	}\label{fig2}
\end{figure}

We next show that our theory confirms computational power of FBM (Fig.~\ref{fig2}). As the training data size increases, the test representation loss [Eq.~\eqref{Hami}] on unseen data pairs decreases first rapidly and then slowly, indicating that the output representation space is well separated given sufficient data size. The separation is well supported by the simulation results in the insets of Fig.~\ref{fig2} (b). In particular, the theory is in an excellent agreement with algorithmic results of BP running on single instances of finite sizes. As expected, the increase of Gaussian variance ($\Delta$) makes the clustering hard, and thus the test generalization loss grows sharply [Fig.~\ref{fig2} (b)]. In this case, multiple layers must be added and learned. 

The fermion-pair distance can be theoretically estimated using the order parameters of our theory. 
\begin{equation}
        D^2_{F} =\int Du_1Du_2 \left[\phi\left( mM + \Delta{\sqrt{q}u_1}\right) - \phi\left( -mM + \Delta{\sqrt{q}u_2}\right)\right]^2.
\end{equation}
This distance is predicted to increase non-linearly with 
$d_F$, but saturates in the large $d_F$ regime. To see the relationship between $d_F$ and the classification accuracy, we define the classification error rate as follows,
\begin{equation}
    \begin{aligned}
        \epsilon_g^C &= \left\langle \Theta\left[-yz\left(\mathbf{x},\mathbf{w}\right)\right]\right\rangle_{\mathbf{w},P(\mathbf{x},y)}
        &=\frac{1}{2}\left[1-\mathrm{erf}\left(\frac{Mm}{\Delta\sqrt{2q}}\right)\right],
    \end{aligned}
\end{equation}
where $\langle\cdot \rangle$ means the average over the test data and the optimal weight configuration following the Boltzmann measure. We can thus get the generalization accuracy as $1-\epsilon^C_g$. Once again, this accuracy is determined by the order parameters. The predicted accuracy increases with $d_F$ until saturation [Fig.~\ref{fig2} (c,d)]. This prediction qualitatively matches the following experimental results on more complex networks and datasets. We next apply FBM to real data learning and show how this framework resolves the conflict between generalization and robustness. In particular, by tuning $d_F$, the FBM can outperform backpropagation when the data size is not sufficient, and even achieve better adversarial robustness.

\subsection{FBM achieves both accuracy and adversarial robustness in practical learning}
We compare FBM with multilayer perceptron (MLP) trained using backpropagation in this section.
The FBM is trained layer by layer with the contrastive cost (Eq.~\eqref{Hami}, but the preactivation of each neuron becomes a vector because of the input data-pair), and only the readout layer is trained by the cross entropy of the classification. In contrast, MLP is trained jointly by a single cost of cross entropy.

\begin{figure}
	\centering
	\includegraphics[width=0.8\textwidth]{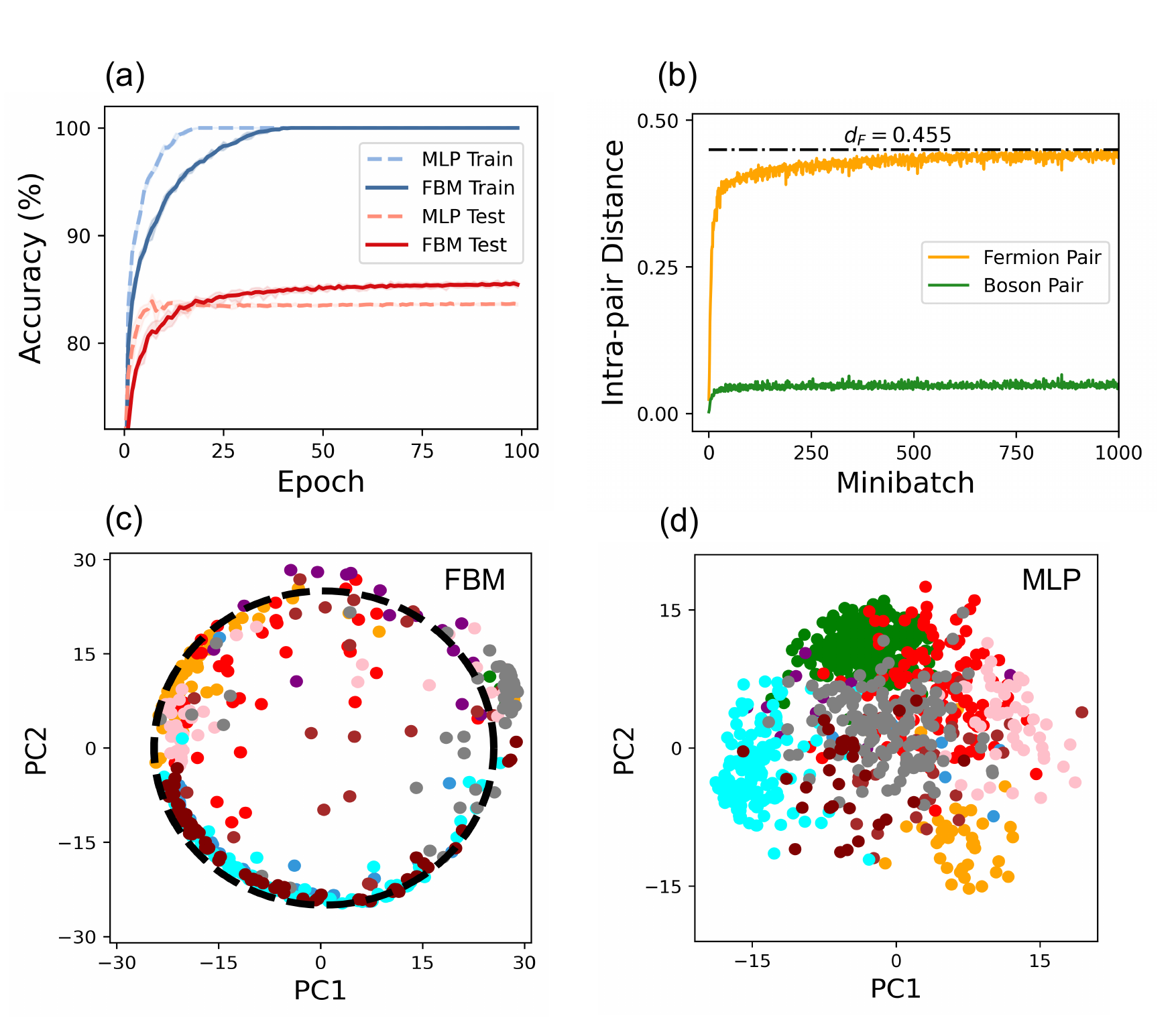}
	\caption{Performance comparison between FBM and MLP. (a) Training and test trajectories on the MNIST dataset~\cite{mnist}. The fluctuation is estimated from five independent runs. The network structure is specified by 784-1000-10, trained by $1000$ handwritten digits and tested by $4000$ digits. $d_F=0.455$. Stochastic gradient descent is used with mini-batch size of $50$. 
        (b) The distance of fermion and boson pairs during training. Other parameters are the same as (a). (c,d) The two-dimensional section of three-dimensional PCA projection of the hidden representation learned by FBM and MLP. The first three eigenvectors of PCA are considered with explained variance ratio $r$.  (c) FBM ($r=74.6\%$). (d) MLP ($r=54.5\%$). Different colors encode different classes. $d_F=1.2$.
        }\label{fig3}
\end{figure}

Compared with MLP, FBM with local learning has better generalization accuracy [Fig.~\ref{fig3} (a)], while the geometry of hidden representation space is well controlled [Fig.~\ref{fig3} (b)]. Interestingly, by applying principal component analysis (PCA) to the hidden representation space, we find that the neural activity is approximately distributed on the surface of a sphere [a section displayed in Fig.~\ref{fig3} (c)] in the case of FBM, while the MLP training has no such intriguing geometric properties [Fig.~\ref{fig3}(d)]. This observation demonstrates that the local contrastive learning leads to low-dimensional disentangled representations, resembling a similar process in visual cortical hierarchy~\cite{Dicarlo-2007,DiCarlo-2012}. FBM thus bears the computational benefit of learning semantically meaningful clusters, and each cluster acts as a prototype of noisy sensory inputs. 

\begin{figure}
	\centering
	\includegraphics[width=0.8\textwidth]{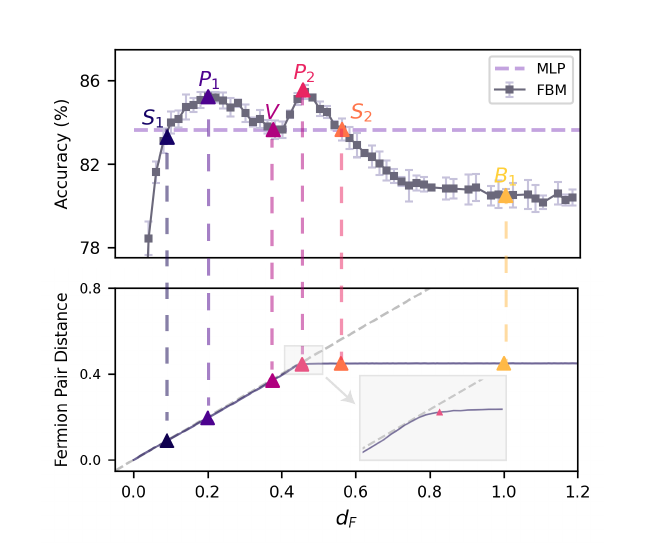}
	\caption{Generalization accuracy tuned by $d_F$ in FBM.  Error bars are the fluctuation with five independent runs. The dash line marks the performance of MLP. The point  $S_1$ marks the target distance where the accuracy of FBM first matches that of MLP; The point $P_1$ marks the first peak; The point $V$ marks the local minimum; The point $P_2$ marks the second peak (the best performance); The point $S_2$ marks the place where the accuracy of FBM drops down to that of MLP; The point $B_1$ marks the stationary part.  The bottom figure shows the fermion-pair distance as a function of $d_F$, where the grey dash line indicates an identity function.}\label{fig4}
\end{figure}

The target distance $d_F$ plays a key role, as predicted by the theory. By varying $d_F$, we find that the generalization accuracy displays double ascent phenomenon [Fig.~\ref{fig4} (a)],  and we also find that the fermion-pair distance first increases approximately linearly with $d_F$ but gets saturated when $d_F>0.455$ [Fig.~\ref{fig4} (b)]. This critical threshold may vary with dataset and architecture details, but this phenomenon is qualitatively the same across different settings (e.g., changing width of hidden layer and training data size), which is also consistent with the theory [Fig.~\ref{fig2}]. The profile of double peaks may depend on real structured dataset. The double ascent phenomenon implies that there exists an optimal representation geometry, which is intimately related to adversarial robustness.  

The standard classifier can be easily fooled by an imperceptible perturbation (adversarial attack). The standard training learns highly predictive yet non-robust features by focusing on spurious correlations between input and output~\cite{Adv-2018,Adv-2018b,Adv-2019,Shortcut-2020}. We argue that a semantically meaningful feature representation should be learned before the classification of last layer. This representation can be formed by our local contrastive learning, resulting in geometrically separated prototype clusters, such that decision boundaries are low density areas. We consider fast gradient sign method (FGSM) and white noise attacks to the trained neural network~\cite{Good-2015,Madry-2018}, i.e., the perturbation to the input $\delta\mathbf{x}=\epsilon\mathrm{sign}(\nabla_{\mathbf{x}}\mathcal{C}) $, and $\epsilon\boldsymbol{\xi}$, respectively. $\mathcal{C}$ is the cross entropy cost, $\epsilon$ is the attack strength, and the white noise is sampled independently as $\xi_i\sim\mathcal{N}(0,1)$.

\begin{figure}
        \centering
        \includegraphics[width=0.8\textwidth]{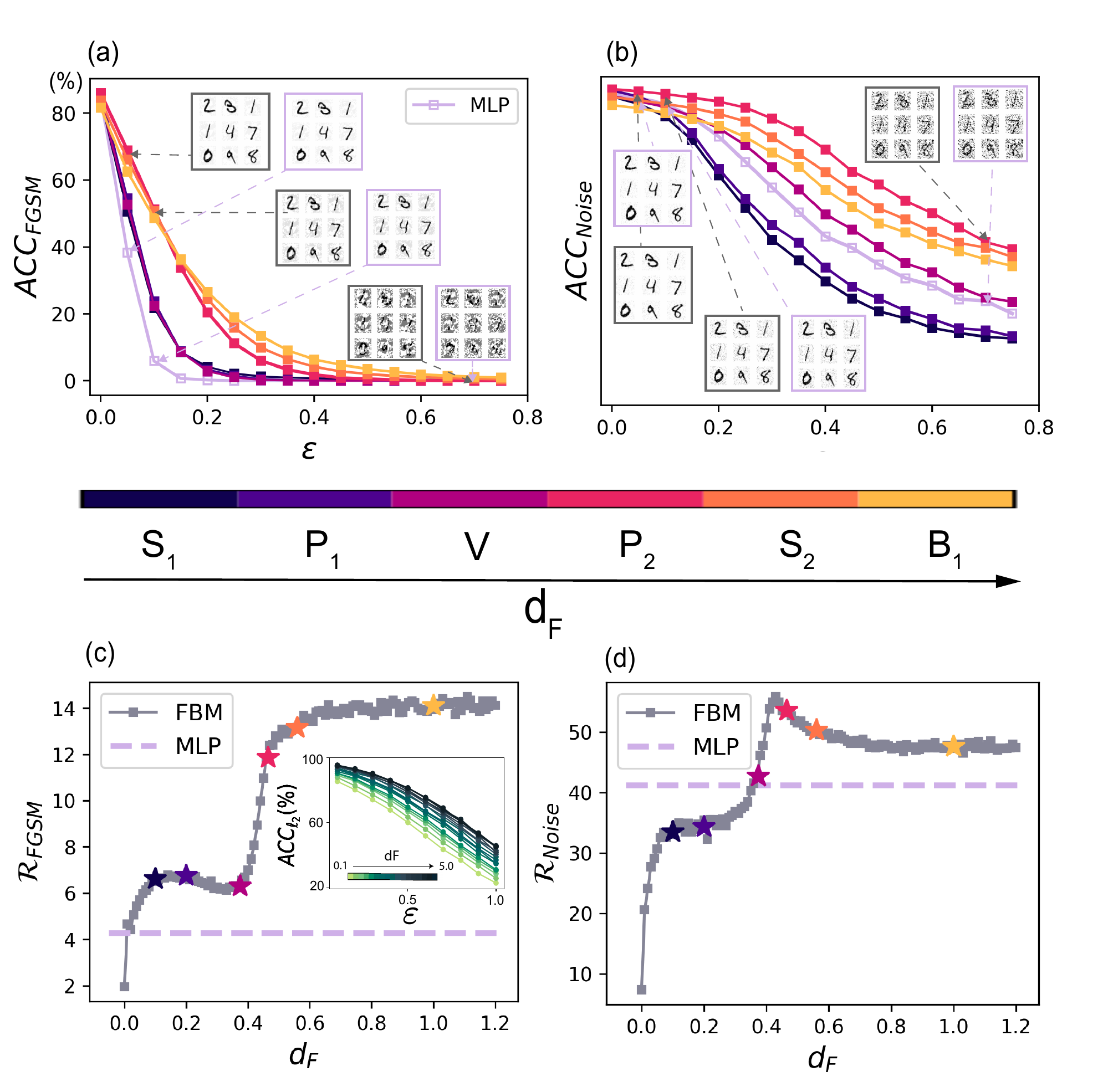}
        \caption{Robustness properties of FBM and MLP. 
        The line color indicates increasing value of $d_F$ corresponding to marked points in Fig.~\ref{fig4}. 
 (a) FGSM attack. (b) Gaussian white noise attack. The insets show adversarially perturbed inputs. (c,d) Adversarial robustness measured by the area under the accuracy-$\epsilon$ curves. The inset in (c) shows a theoretical prediction of $\ell_2$ norm attack.}
        \label{fig5}
    \end{figure}

The adversarial robustness of FBM can be explained by an $\ell_2$ norm attack (FGSM is $\ell_\infty$ norm bounded attack) defined below,
\begin{equation}
    \begin{aligned}
        \delta \mathbf{x} = {\epsilon} \frac{\nabla_{\mathbf{x}}\mathcal{L}}{\sqrt{\sum_i\left(\nabla_{{x_i}}\mathcal{L}\right)^2}},
    \end{aligned}
\end{equation}
where we consider a mean-squared loss $\mathcal{L}=\frac{1}{2}\left[y-\tanh(\sum_iw_ix_i)\right]^2$. Based on the mean-field theory derived before,
 the averaged adversarial accuracy as a function of attack strength is given by
\begin{equation}
    \begin{aligned}
   & {\rm ACC}_{adv} = 1 - \left\langle \Theta\left[-y\left(z + \mathbf{w}^\t\delta\mathbf{x}\right)\right]\right\rangle_{\mathbf{w},x,y}\\
        &=\int Du\Theta\left(mM +\Delta\sqrt{q} u +\epsilon\sqrt{q}\mathrm{sgn}\left[\mathcal{K}_1\left(mM +\Delta\sqrt{q} u\right)\right]\right),
      \end{aligned}
\end{equation}
where $\mathcal{K}_y\left(z\right) = -\left[y-\phi\left(z\right)\right]\phi^{\prime}\left(z\right)$.
Detailed derivations are given in Methods. As predicted by the above theory [inset of Fig.~\ref{fig5} (c)], $d_F$ is a key parameter to control the adversarial robustness. In a narrow range of $d_F$, the robustness measured by the area under the accuracy-$\epsilon$ curve (Fig.~\ref{fig5}) increases sharply for both types of attack. For FGSM, the robustness gets saturated while the Gaussian noise attack exhibits a peak. This demonstrates that FBM without adversarial training is more adversarial robust than MLP, providing a principled route toward alignment with human cognition processing using disentangled invariant representations~\cite{Dicarlo-2007,DiCarlo-2012,Huang-2018,Zhou-2021}.

\section{Concluding remarks}
A fundamental question in machine learning and neuroscience is what is a good representation learning that is biological plausible and moreover adversarial robust. We propose a local contrastive learning, namely Fermi-Bose machine to resolve this challenging question. Our theoretical toy model based on Gaussian mixture data captures main characteristics of FBM, explaining effects of data density, entanglement intensity and target fermion-pair distance. When applied to MNIST data learning, FBM exhibits intriguing geometric separation, double ascent of accuracy and a great improvement of adversarial robustness over joint training of MLP.

Impacts on interdisciplinary fields across machine learning, neuroscience and physics are direct. First, without adversarial training, the local contrastive learning achieves both high standard accuracy and adversarial robustness. Second, the local contrastive learning reveals how abstract concepts are formed by geometric separation, aligning with cognitive information processing in cortex. Third, working mechanisms of Fermi-Bose machine are transparent, as the learning can be formulated by a data-pair dependent partition function, which can be solved by mean-field theory.
Many future directions arise, such as connection to supervised Hebbian learning~\cite{Barra-2023}, precise mathematical mechanism of robustness versus representation geometry, generalization to unsupervised learning and so on.

\section{Acknowledgments}
H.H. benefited from the international workshop of machine learning physics held in Kyoto University (Nov 16-18, 2023). This research was supported by the National Natural Science Foundation of China for
Grant number 12122515, and Guangdong Provincial Key Laboratory of Magnetoelectric Physics and Devices (No. 2022B1212010008), and Guangdong Basic and Applied Basic Research Foundation (Grant No. 2023B1515040023).

\onecolumngrid
\appendix
\section{Algorithmic details of FBM training}
The deep network we consider consists of $L$ layers, with $L-2$ hidden layers. Because the network is trained layer by layer,  we consider training the weights between $\ell^{th}$ and $(\ell+1)^{th}$ layers as an example to detail the training method. The $\ell^{th}$ layer of the network has $N_\ell$ neurons. The connection weight between the neuron $i$ in the $(\ell+1)^{th}$ layer and the neuron $j$ in the $\ell^{th}$ layer is denoted by $w_{ij}^{\ell+1}$.

Firstly, we introduce how to construct fermion or boson pairs for training.  The pairing process is exhaustive and does not miss any possible pairs. For instance, we have a total of 10 images in a minibatch, consisting of two images of handwritten digit zero and eight images of handwritten digit one.  We thus have sixteen fermion pairs and twenty-nine  boson pairs. Subsequently, we assign pair-labels to boson pairs as ($+1$) and to fermion pairs as ($-1$). This label setting is consistent with that introduced in the main text. Note that this label is not the original label of digit images, but used to define the cost function below. We remark that this pairing process is akin to data augmentation, especially useful when the training data is scarce. 

For the sake of compactness, we define the pre-activation in the form of vector:
    \begin{equation}
        z^\ell_j ={\begin{pmatrix}
                    z^\ell_{L,j} \\
                    z^\ell_{R,j}
                    \end{pmatrix}},
    \end{equation}
where we use $L$ and $R$ to distinguish which element of the data pair triggers the network output. In the standard training of MLP, $z^\ell_j$ is a scalar quantity, but in FBM, it is a vector as defined above.
The pre-activations in the next layer can be estimated according to the following affine transformation:
   \begin{equation}
        z ^{\ell+1}_j
            = \sum_j w_{ij}^{\ell+1}
                    {\begin{pmatrix}
                    h^{\ell}_{L,j} \\
                    h^{\ell}_{R,j}
                    \end{pmatrix}},
    \end{equation}
where the activation $\mathbf{h}^\ell_j$ is also a vector whose components $h^\ell_{X,j}=\phi(z^\ell_{X,j}$) ($X=L,R$), and in practice we use $\phi(x)=\frac{1}{2}[\tanh(x)+1]$, but other choices are also possible, and we would not discuss the effects of transfer function here.

The loss function to minimize for each data pair is expressed below:
    \begin{equation}\label{loss}
        \mathcal{L} = \frac{1 + \sigma}{4} D^2+\frac{1-\sigma}{4} \varphi\left(d_F - D ^2\right)+\lambda_w\Vert\mathbf{w}\Vert^2,
    \end{equation}
where the pair label $\sigma\in \{-1,1\}$, and $D^2 =\Vert \mathbf{h}^{\ell+1}_L-\mathbf{h}^{\ell+1}_R\Vert^2_2$ ($\ell$ could be a value taken from
$1$ to $L-1$). Note that this is a layerwise training. The first term of boson pair distance should be minimized, and the second term is related to the fermion pair distance, which must reach the target separation $d_F$. $\varphi(\cdot)$ is some relu-like function. 
The loss function is layerwisely minimized by stochastic gradient descent (update rules of weights are detailed below). The full training data is first divided into
mini-batches each of which contains a few randomly sampled images. These images inside one mini-batches are paired according the aforementioned rule.
 After a layer is trained, the weights before this layer
are all frozen, and then the next layer is trained following a similar procedure. All layers except the last readout layer are trained by minimizing the loss function in 
in Eq.~\eqref{loss}. The readout layer is trained by minimizing a cross-entropy cost function for classification we consider in this paper. The test accuracy is estimated 
from a test data set which is fed into the trained network one by one. Hyper-parameters for the network and algorithm are summarized in the Table~\ref{tab}.

  \begin{table}[h]
        \centering
        \begin{tabular}{l l l}
            \hline
            Parameter & Value & Description \\
            \hline
            $N_0$ & 784 & input dimension \\
            $N_1$ & 1000 & hidden layer width \\
            $N_3$ & 10 & the number of output units \\
            $M$ & 1000 & training dataset size  \\
            $\mathcal{B}$ & 50 & mini-batch size \\
            $\lambda_w$ & 0.01 & weight decay strength \\
            \hline
        \end{tabular}
        \caption{Hyperparameters used in training FBM to learn MNIST. }
        \label{tab}
    \end{table}

Next, we detail how the stochastic gradient descent works, which is divided into two coupled steps. The first forward propagation step is summarized 
as follows:
\begin{equation}
        \begin{aligned}
        \mathbf{z} ^\ell_j &={\begin{pmatrix}
                    z^\ell_{L,j} \\
                    z^\ell_{R,j}
                    \end{pmatrix}}
                    =\sum_k w_{jk}^\ell {\begin{pmatrix}
                   h^{\ell-1}_{L,k} \\
                 h^{\ell-1}_{R,k}
                    \end{pmatrix}},\\
        \mathbf{h} ^\ell_j 
                    &=\phi{\begin{pmatrix}
                    z^\ell_{L,j} \\
                    z^\ell_{R,j}
                    \end{pmatrix}}.
        \end{aligned}                
    \end{equation}
The second backward propagation proceeds as follows:
\begin{equation}\label{back}
        \begin{aligned}
        \Delta^\ell_j 
                    &\equiv \frac{\partial \mathcal{L}}{\partial \mathbf{z}^\ell_j}
                    ={\begin{pmatrix}
                    \frac{\partial \mathcal{L}}{\partial z_{L,j}^\ell}\\
                    \frac{\partial \mathcal{L}}{\partial z_{R,j}^\ell}
                    \end{pmatrix}}
                    ={\begin{pmatrix}
                    \frac{\partial \mathcal{L}}{\partial h_{L,j}^\ell}\\
                    \frac{\partial \mathcal{L}}{\partial h_{R,j}^\ell}
                    \end{pmatrix}}
                    \odot 
                    {\begin{pmatrix}
                    \frac{\partial h_{L,j}^\ell}{\partial z_{L,j}^\ell}\\
                    \frac{\partial h_{R,j}^\ell}{\partial z_{R,j}^\ell}
                    \end{pmatrix}}
                    ={\begin{pmatrix}
                    \frac{\partial \mathcal{L}}{\partial h_{L,j}^\ell}\\
                    \frac{\partial \mathcal{L}}{\partial h_{R,j}^\ell}
                    \end{pmatrix}}
                    \odot 
                    {\begin{pmatrix}
                   \phi'(z_{L,j}^\ell)\\
                    \phi'( z_{R,j}^\ell)
                    \end{pmatrix}},\\
       \delta^\ell_{jk}
                    &\equiv\frac{\partial \mathcal{L}}{\partial w_{jk}^{\ell}} 
                    =\frac{\partial \mathcal{L}}{\partial \mathbf{z}^\ell_j}\frac{\partial \mathbf{z}^\ell_j} {\partial w_{jk}^{\ell}}
                    = {\left ({\mathbf{h}^{\ell-1}_k} \right ) }^{\operatorname{T}} \Delta^\ell_j,
\end{aligned}
\end{equation}
where $\odot$ represents the element-wise product. We remark here that this is just a gradient descent without propagating a global error from the top layer, thereby unlike backpropagation in standard deep learning. The derivative $\frac{\partial \mathcal{L}}{\partial\mathbf{h}^\ell_j}$ in Eq.~\eqref{back} can be directly computed using the definition of  the local contrastive loss $\mathcal{L}$.
The weight is then updated as follows:
  \begin{equation}
        w^\ell_{jk} \leftarrow w^\ell_{jk}-\eta\delta^\ell_{jk},
        \end{equation}
where $\eta$ is the learning rate, set by default values of the Adam algorithm~\cite{Adam}. Codes realizing the FBM training are available in our GitHub page~\cite{MS-2024} (to be released upon formal publication of the paper), which also includes 
two kinds of adversarial attacks.

\section{Theoretical details of replica calculation}
\subsection{Finite temperature}
To get a quantitative description of local contrastive learning, we semi-rigorous analyze FBM in a simple setting of $N$-$1$ structure, as already detailed in the main text. This statistical mechanics analysis can be carried out by application of replica method, a seminal method in disordered system theory~\cite{Mezard-1987}. The disorder in our model comes from the design of Gaussian mixture data, which is detailed as follows.
	
The training data is given by $\{\left(\mathbf{x}^{\mu}_1, {y_1}^\mu\right), \left(\mathbf{x}^{\mu}_2,{y_2}^\mu\right)\}_{\mu\in[P]}$,  where $\mathbf{x}^\mu_1,\mathbf{x}^\mu_2 \in \mathbb{R}^N$,  and $y_1^\mu$ or $ y_2^\mu$ is a scalar. Each $\mu$ is an input data-pair, which can be a boson or fermion pair.  The statistics of each pair $\mathrm{P}\left(\mathbf{x}^{\mu}_1, y_1^\mu, \mathbf{x}^{\mu}_2, y_2^\mu \right)= \mathrm{P}\left(y^\mu_1,  y^\mu_2 \right)\mathrm{P}\left(\mathbf{x}_1^{\mu}\vert y_1^\mu\right)\mathrm{P}\left(\mathbf{x}_2^{\mu}\vert y_2^\mu\right)$, where
	\begin{equation}
		\begin{aligned}
&\mathrm{P}\left(\mathbf{x}_1^{\mu}\vert y_1^\mu\right) = \mathcal{N}\left(y_1^\mu \id_N \frac{m}{N}, \frac{\Delta^2}{N}\mathbf{I}_N\right),\\
&\mathrm{P}\left(\mathbf{x}^{\mu}_2\vert y_2^\mu\right) = \mathcal{N}\left(y_2^\mu \id_N \frac{m}{N},\frac{\Delta^2}{N}\mathbf{I}_N\right),\\
&\mathrm{P}\left(y_2^{\mu}, y_1^\mu\right) = \mathrm{P}\left(y_1^\mu \right)\mathrm{P}\left(y_2^\mu\vert y_1^\mu\right) =\frac{1}{2}\rho\delta(y_1^\mu-y_2^\mu)+  \frac{1}{2}\left(1 - \rho\right)\delta (y_1^\mu +y_2^\mu),
		\end{aligned}
	\end{equation}
	where $\delta(\cdot)$ is a Dirac delta function,  $\id_N$ is an all-one vector,  and $\mathbf{I}_N \in \mathbb{R}^{N\times N}$ is an identity matrix.
	
  The single-node output is given by $h^\mu =\phi(z^\mu)=\phi\left(\sum_{i}w_{i}x^\mu_{i}\right)$. The loss is given below:
  \begin{equation}\label{Hami-app}
    \begin{aligned}
      \mathcal{L} =\sum_{\mu=1}^P\frac{1}{2} \left[\sigma^\mu D_\mu^2+(1-\sigma^\mu) \varphi\left(d_F - D_\mu ^2\right) \right]+\frac{\lambda_w}{2}\Vert\mathbf{w}\Vert^2_2,
      \end{aligned}
    \end{equation}
where $\lambda_w$ is a weight regularization parameter commonly used in deep learning, $\sigma^\mu=\frac{1+y_1^\mu y_2^\mu}{2}$, $y_1^\mu$ and $y_2^\mu$ are the labels of $\mathbf{x}^\mu_1$ and $\mathbf{x}^\mu_2$, respectively, and $D_\mu^2 = [\phi(z_1^\mu)-\phi(z_2^\mu)]^2$ is the Euclidean distance. $z_1 (z_2)$ is the preactivation for the first (second) element of the data pair input. $\varphi(\cdot)$ denotes a relu-like function, and we choose $\varphi(x) =\frac{x + \sqrt{x^2 + a}}{2}$ $(a=0.01)$ for function smoothness at the origin point~\cite{Barron-2021}.
	 
The partition function for our model reads,
	\begin{equation}
		Z= \int \prod^{N}_{i=1}\mathrm{d}w_{i}\mathrm{P}_0(w_i) \mathrm{exp}\left[-\beta \sum^P_{\mu=1}\mathcal{L}^\mu\left(\mathbf{w}\right)\right],
	\end{equation}	
	where $\mathcal{L}^\mu\equiv\frac{1}{2} \left[\sigma^\mu D_\mu^2+(1-\sigma^\mu) \varphi\left(d_F - D_\mu ^2\right) \right]$,
	 $\mathbf{w} \in \mathbb{R}^{N}$,  $\mathrm{P}_0(w_{i}) = \mathrm{e}^{-\frac{\beta{\lambda_w}}{2} w^2_{i}}$ ($\forall i$) corresponding to weight regularization commonly used in deep learning. In our case, the partition function is a random variable, resulting in a random free energy in physics. We must do an average of the free energy over Gaussian mixture data. We have thus to use replica method as follows: 
	 	\begin{equation}
		-\beta f = \frac{\mathop{\mathbb{E}}\limits_{\mathbf{X}_1,\mathbf{y}_1, \mathbf{X}_2,\mathbf{y}_2}\mathrm{ln}Z}{N}=\mathrm{lim}_{n\to 0}\frac{\mathrm{ln}\left(\mathop{\mathbb{E}}\limits_{\mathbf{X}_1,\mathbf{y}_1, \mathbf{X}_2,\mathbf{y}_2} Z^n\right)}{Nn},
	\end{equation}	
	where $f$ is free energy per synapse (density), $n$ is a replica number first taken to an integer and finally sent to zero (a bit weird but exact in some cases~\cite{HH-2022}), $\mathbf{X}_{1} \in \mathbb{R}^{N\times P}$ whose $\mu$-th column $\mathbf{X}_1^\mu=\mathbf{x}_1^\mu$ ($\mathbf{X}_2$ has a similar meaning),  and $\mathbf{y} \in \mathbb{R}^P$ (the subscript $1$ (or $2$) indicates which element of an input data-pair). In the following, we use $\mathbb{E}$ to represent the average over the data disorder.

Now, we explicitly write down the integer power of the random partition function averaged over the data pairs.
\begin{equation}
		\begin{aligned}
			\mathbb{E}Z^n &= \sum_{\{\mathbf{y}_1,\mathbf{y}_2\}}\int  \prod_\mu\Bigl[\mathrm{d}\mathbf{x}_1^\mu\mathrm{d}\mathbf{x}_2^\mu\Bigr] \prod_\mu \left[\mathrm{P}(y_1^\mu,y_2^\mu)\prod_{i}\left[\mathrm{P}\left(\left(\mathbf{x}_1^\mu\right)_i|y_1^\mu\right)\mathrm{P}\left(\left(\mathbf{x}_2^\mu\right)_i|y_2^\mu\right)\right]\right]\int \prod^{n}_{a=1}\mathrm{d}\mathbf{w}^a \mathrm{P}_0(\mathbf{w}^a)\\
			&\times\prod_{a=1}^{n}\prod_{\mu}\mathrm{exp}\left\{-\beta\left[\frac{\sigma^\mu}{2}\left(D^{a}_\mu\right)^2+\frac{1-\sigma^\mu}{2} \varphi\left[d_F - \left(D^{a}_\mu\right)^2\right]\right] \right\},
		\end{aligned}
	\end{equation}
	where $D^{a}_\mu = \sqrt{\left[\phi\left(\mathbf{w}^a\mathbf{x}_1^\mu\right)-\phi\left(\mathbf{w}^a\mathbf{x}_2^\mu\right)\right]^2}$,  and $a$ is the index of replicas ($a = 1,\ldots,n$).	
	
	To further simplify the calculation, we define the following conditional probability distributions:
	\begin{equation}
		\begin{aligned}
		&\mathrm{P}\left({z}_1^{a\mu}|y_1^{\mu}\right) =  \langle\delta\left(z_1^{a\mu}-\mathbf{w}^a\mathbf{x}_1^{\mu}\right)\rangle_{\mathbf{x}_1^{\mu}|y_1^{\mu}},\\
			&\mathrm{P}\left(z_2^{a\mu}|y_2^{\mu}\right) =  \langle\delta\left(z_2^{a\mu}-\mathbf{w}^a\mathbf{x}_2^{\mu}\right)\rangle_{\mathbf{x}_2^{\mu}|y_2^{\mu}}.
					\end{aligned}
	\end{equation}
After a straightforward computation, we get Gaussian distributions for the preactivations as follows:
\begin{equation}
		\begin{aligned}
			\mathrm{P}(\mathbf{z}_1|\mathbf{y}_1)&=\prod_\mu \mathrm{P}(\mathbf{z}_1^{\mu}|y_1^{\mu})\\
			&=\frac{1}{\sqrt{(2\pi)^{nP}(\det{\Delta^2 \mathbf{Q}})^P}}\prod_\mu \mathrm{exp}\left[\left(\mathbf{z}_1^{\mu}- my_1^\mu \mathbf{M}  \right)^\t{(2\Delta^2\mathbf{Q})^{-1}} \left(\mathbf{z}_1^{\mu}- my_1^\mu \mathbf{M}\right) \right],
		\end{aligned}
	\end{equation}	
and
\begin{equation}
		\begin{aligned}
			\mathrm{P}(\mathbf{z}_2|\mathbf{y}_2)&=\prod_\mu \mathrm{P}(\mathbf{z}_2^{\mu}|y_2^{\mu})\\
			&=\frac{1}{\sqrt{(2\pi)^{nP}(\det{\Delta^2 \mathbf{Q}})^P}}\prod_\mu \mathrm{exp}\left[\left(\mathbf{z}_2^{\mu}- my_2^\mu \mathbf{M}  \right)^\t{(2\Delta^2\mathbf{Q})^{-1}} \left(\mathbf{z}_2^{\mu}- my_2^\mu \mathbf{M}\right) \right],
		\end{aligned}
	\end{equation}	
		where the preactivations $\mathbf{z}_1^\mu, \mathbf{z}_2^\mu \in \mathbb{R}^{n}$, and
	 $\mathbf{M} \in \mathbb{R}^{n}, \mathbf{Q} \in \mathbb{R}^{n \times n}$ are mean and correlation of weights in the replica space.
	 Their more precise expressions are given below. 
	\begin{equation}
		\begin{aligned}
			&\mathbf{M} = \begin{pmatrix}
			        \frac{\sum_iw_i^1}{N}\\
				\vdots\\
				\frac{\sum_i{w}^a_i}{N}\\
				\vdots\\
				\frac{\sum_i{w}^n_i}{N}
			\end{pmatrix},
			\quad 
			\mathbf{Q} = \begin{pmatrix}
				\frac{\sum_i{w}^1_i  {w}^1_i}{N}&\cdots&\frac{\sum_i{w}^1_i {w}^n_i}{N}\\
				\vdots&\ddots&\vdots\\
				\frac{\sum_i{w}^n_i  {w}^1_i}{N}&\cdots&\frac{\sum_i{w}^n_i {w}^n_i}{N}
			\end{pmatrix}.
		\end{aligned}
	\end{equation}

One can verify $\mathbf{M}$ and $\mathbf{Q}$ are related to the mean and variance of the above
 Gaussian distribution of preactivations. This is demonstrated below.
\begin{equation}\label{covz}
		\begin{aligned}
			&\langle \mathbf{z}_1^\mu \rangle_{\mathbf{x}_1^\mu|{y_1}^\mu} = my_1^\mu\mathbf{M},\\
			&\langle \mathbf{z}_2^\mu \rangle_{\mathbf{x}_2^\mu|{y_2}^\mu} = my_2^\mu\mathbf{M},\\
			&\langle \mathbf{z}_1^\mu\left(\mathbf{z}_1^\nu\right)^\t \rangle_{\mathbf{x}_1^\mu| y_1^\mu} - \langle \mathbf{z}_1^\mu \rangle_{\mathbf{x}_1^\mu| y_1^\mu}\langle \mathbf{z}_1^\nu \rangle_{\mathbf{x}_1^\mu|y_1^\mu}^\t = \Delta^2\mathbf{Q}\delta_{\mu\nu},\\
			&\langle \mathbf{z}_2^\mu\left(\mathbf{z}_2^\nu\right)^\t \rangle_{\mathbf{x}_2^\mu|y_2^\mu} - \langle \mathbf{z}_2^\mu \rangle_{\mathbf{x}_2^\mu|y_2^\mu}\langle \mathbf{z}_2^\nu \rangle_{\mathbf{x}_2^\mu|y_2^\mu}^\t = \Delta^2\mathbf{Q}\delta_{\mu\nu},\\
			&\langle \mathbf{z}_2^\mu\left(\mathbf{z}_1^\nu\right)^\t \rangle_{\mathbf{x}_2^\mu|y_2^\mu, \mathbf{x}_1^\mu|y_1^\mu} - \langle \mathbf{z}_2^\mu \rangle_{\mathbf{x}_2^\mu|y_2^\mu}\langle \mathbf{z}_1^\nu \rangle_{\mathbf{x}_1^\mu|y_1^\mu}^\t = 0,
		\end{aligned}
	\end{equation}
where the last equality is derived due to the fact that given the label $y$, the data sample $\mathbf{x}$ is independently generated. $\mathbf{M}$ and $\mathbf{Q}$ can then be used to represent the Gaussian distribution, which would greatly simplify the replica calculation. Hence, we can insert Dirac delta functions to specify $\mathbf{M}$ and $\mathbf{Q}$, and then use their Fourier representations:
\begin{equation}
		\begin{aligned}
			& \prod^n_a\delta\left(N{M}^{a} - {\sum_i{w}^a_i}\right)=  \int^{+i\infty}_{-i\infty} \prod^n_a \frac{\operatorname{d}{\hat{M}}^{a}}{\left(2\pi i\right)} \mathrm{exp}\left[ {\hat{M}}^{a} \left(N{M}^{a} - \sum_i{w}^a_i \right)\right],\\
			& \prod_{a} \delta\left(N{Q}^{aa} - {\sum_i{w}^a_i{w}^a_i}\right)=  \int^{+i\infty}_{-i\infty} \prod_{a} \frac{\operatorname{d}{\hat{Q}}^{aa}}{\left(2\pi i\right)} \mathrm{exp}\left[{\hat{Q}}^{aa}\left(N{Q}^{aa} - \sum_i{w}^a_i {w}^a_i\right)\right],\\
			& \prod_{a< b} \delta\left(N{Q}^{ab} - {\sum_i{w}^a_i{w}^b_i}\right)=  \int^{+i\infty}_{-i\infty} \prod_{a< b} \frac{\operatorname{d}{\hat{Q}}^{ab}}{\left(2\pi i\right)} \mathrm{exp}\left[{\hat{Q}}^{ab}\left(N{Q}^{ab} - \sum_i{w}^a_i{w}^b_i\right)\right].		
		\end{aligned}
	\end{equation}
Therefore,
\begin{equation}
		\begin{aligned}
			\mathbb{E} Z^n &=\int^{+i\infty}_{-i\infty} \left(\prod_{a\leq b} \frac{\operatorname{d}\hat{{Q}}^{ab}\operatorname{d}{Q}^{ab}}{\left(2\pi i\right)}\right)\left(\prod^n_{a} \frac{\operatorname{d}\hat{{M}^{a}}\operatorname{d}{M}^{a}}{\left(2\pi i\right)}\right)\\
			&\times \int \prod^n_{a} \operatorname{d}\mathbf{w}^a \prod^n_{a}\operatorname{P}_0\left(\mathbf{w}^a\right)\mathrm{exp}\left(N\sum_{a < b}\left({\hat{{Q}}}^{ab}{Q}^{ab}\right) -\sum_{a<b}\hat{{Q}}^{ab}\sum_i{w}^a_i{w}^{b}_i\right)\\
			&\times \mathrm{exp}\left(N\sum^n_{a}{\hat{M}}^{a}{M}^{a} - \sum^n_{a}{\hat{M}}^{a}\sum_i{w}^a_i \right)\mathrm{exp}\left(N\sum\limits_{a}^n{\hat{{Q}}}^{aa}{Q}^{aa} -\sum\limits_{a}^n\hat{{Q}}^{aa}\sum_i{w}^a_i{w}^{a}_i\right)\\
			&\times  \prod^P_\mu \sum_{\{y_1^\mu,y_2^\mu\}}\int \mathrm{d}\mathbf{z}_1^\mu\mathrm{d}\mathbf{z}_2^\mu \mathrm{P}(y_1^{\mu},y_2^{\mu})\mathrm{P}(\mathbf{z}_2^{\mu}|y_2^{\mu})\mathrm{P}(\mathbf{z}_1^{\mu}|y_1^{\mu})\\
			&\times\prod_{\mu, a}\operatorname{exp}\left[-\beta\left(\frac{\sigma^\mu}{2}\left(D^{a}_\mu\right)^2+\frac{1-\sigma^\mu}{2} \varphi\left[d_F - \left(D^{a}_\mu\right)^2\right] \right)\right].
		\end{aligned}
	\end{equation}

Finally, the integer power of the partition function averaged over data can be recast into a mathematically compact form:
\begin{equation}
		\begin{aligned}
			\mathbb{E} Z^n &=\int \left(\prod_{a\leq b} \frac{\operatorname{d}\hat{{Q}}^{ab}\operatorname{d}{Q}^{ab}}{2\pi i}\right)\left(\prod_{a} \frac{\operatorname{d}\hat{{M}^{a}}\operatorname{d}{M}^{a}}{2\pi i}\right)e^{-N\beta f},\\
		\end{aligned}
	\end{equation}
	where $-\beta f$ is an action in physics and can be expressed as the sum of the following three terms:
	\begin{equation}
		\begin{aligned}
			&\Psi_{0} = \sum^n_a{\hat{M}}^{a}{M}^a +\sum^n_{a}{\hat{{Q}}}^{aa}{Q}^{aa} + \sum_{a< b}{\hat{{Q}}}^{ab}{Q}^{ab},\\
			&\Psi_{S}  = \frac{1}{N}\operatorname{ln}\int \prod^n_a \operatorname{d}\mathbf{w}^a \prod^n_a\operatorname{P}_0\left(\mathbf{w}^a\right) \operatorname{exp}\left(-\sum\limits_a^n\hat{{Q}}^{aa}\sum_i{w}^a_i{w}^{a}_i -\sum\limits_{a<b}^{n\left(n-1\right)/2}\hat{{Q}}^{ab}\sum_i{w}^a_i{w}^{b}_i-\sum_a^n\hat{{M}}^{a}{w}^a_i\right),\\
			&\Psi_{E} =   \alpha \operatorname{ln}\sum_{\{y_1,y_2\}}\int \mathrm{d}\mathbf{z}_1\mathrm{d}\mathbf{z}_2\mathrm{P}(y_1, y_2 )\mathrm{P}(\mathbf{z}_2 \vert y_2 )\mathrm{P}(\mathbf{z}_1\vert y_1)\prod^{n}_{a}\operatorname{exp}\left[-\beta\left(\frac{\sigma}{2}\left(D^{a}\right)^2+\frac{1-\sigma }{2} \varphi\left[d_F - \left(D^{a}\right)^2\right] \right)\right],
		\end{aligned}
	\end{equation}	
where $(D^a)^2 = [\phi(z_1^a)-\phi(z_2^a)]^2$, and $\sigma=\frac{1+y_1y_2}{2}$.

To proceed, we have to assume the replica symmetry (RS) ansatz specified as follows:
\begin{equation}
		\begin{aligned}
			&{\hat{M}}^a = {\hat{M}}, \quad {M}^a = {M}\quad\forall a,\\
			&{\hat{Q}}^{aa} = {\hat{q}},\quad {Q}^{aa} = {q}\quad\forall a, \\
			&{\hat{Q}}^{ab} = {\hat{Q}}, \quad {Q}^{ab} = {Q} \quad \forall a \not= b.
		\end{aligned}
	\end{equation}	
In essence, RS assumes permutation symmetry of the replica overlap matrix, which is the first level of approximation that should be cross-checked by numerical
experimental results. If this ansatz is not correct, the resulting saddle-point equation (SDE, see below) does not converge during iterations. Then an advanced level of approximation is required~\cite{HH-2022}.
Under the RS ansatz, the replica overlap matrices can be expressed as follows:
\begin{equation}
		\begin{aligned}
			&\mathbf{M} = \begin{pmatrix}
			M\\
				\vdots\\
				{M}\\
				\vdots\\
				{M}
			\end{pmatrix},
			\quad 
			\mathbf{Q} = \begin{pmatrix}
				{q}&{Q}&\cdots&{Q}\\
				{Q}&{q}&\cdots&{Q}\\
				\vdots&\vdots&\ddots&\vdots\\
				{Q}&&\cdots&{q}
			\end{pmatrix}.
		\end{aligned}
	\end{equation}

Applying the RS ansatz to the entropy term $\Psi_S$, we obtain
\begin{equation}
		\begin{aligned}
			\Psi_{S}&= \frac{1}{N}\operatorname{ln}\int \prod^n_a \operatorname{d}\mathbf{w}^a \prod^n_a\operatorname{P}_0\left(\mathbf{w}^a\right) \operatorname{exp}\left(-\sum\limits_a^n\hat{{q}}\sum_i^N{w}^a_i{w}^{a}_i -\sum_{a<b}\hat{{Q}}\sum_i^N{w}^a_i{w}^{b}_i -\sum_a^n\hat{{M}}\sum_i^N{w}^a\right)\\
			&= \frac{1}{N}\operatorname{ln}\int \prod^n_a \operatorname{d}\mathbf{w}^a \prod^n_a\operatorname{P}_0\left(\mathbf{w}^a\right) \operatorname{exp}\left(-\frac{\sum\limits_a^n\left(2\hat{{q}} - \hat{{Q}}\right)\sum_i{w}^a_i{w}^{a}_i}{2}-\frac{\hat{Q}}{2}\sum\limits_{i}^{N}\left(\sum_a^n{w}^a_i\right)^2-\sum_a^n\hat{{M}}\sum_i{w}^a_i\right)\\
			&= \frac{1}{N}\operatorname{ln}\int \prod_{i}D\xi_i \int \prod^n_{a,i} \operatorname{d}{w}^a_i \prod_{a,i}\operatorname{P}_0\left({w}^a_i\right)\\
			&\times \operatorname{exp}\left(-\frac{\sum\limits_a^n\left(2\hat{{q}} - \hat{{Q}}\right)\sum^N_i{w}^a_i{w}^{a}_i }{2}+\sum^N_i\left({\xi}_i\sqrt{-\hat{{Q}}} \sum\limits_{a}^{n}{w}^a_i \right)-\sum_a^n\hat{{M}}\sum^N_i{w}^a_i\right)\\
			&= \operatorname{ln}\int D\xi \int \prod^n_a \operatorname{d}w^a \operatorname{exp}\left(-\frac{\sum\limits_a^n\left(2\hat{{q}}- \hat{{Q}}+\beta\lambda_w\right)({w}^a)^2}{2}+\left({\xi}\sqrt{-\hat{Q}} \sum\limits_{a}^{n}{w}^a \right)-\sum_a^n\hat{M}{w}^a\right),\\ 
		\end{aligned}
	\end{equation}
where $D\boldsymbol{\xi}\equiv\prod_iD\xi_i$ is a Gaussian measure, and the integral identity $\int D\xi e^{b\xi}=e^{b^2/2}$ is used.	

Next we take the limit of $n\to0$, and obtain the limit of $\Psi_S(n)/n$ as the new $\Psi_S$ which reads,
	\begin{equation}
		\begin{aligned}
			\Psi_{S}
			& =\lim_{n\to 0}\frac{1}{n}\ln\int D\xi\left[\int \mathrm{d}w\exp\left( -\frac{\left(2\hat{q}-\hat{Q}+\beta\lambda_w\right)w^2}{2} + \left(\xi\sqrt{-\hat{Q}} -\hat{M}\right)w \right)\right]^n\\
			&= \lim_{n\to 0}\frac{1}{n}\ln\int D\xi\left[\sqrt{\frac{2\pi}{2\hat{q}-\hat{Q}+ \beta\lambda_w}}\exp\left[\left(\xi\sqrt{-\hat{Q}} - \hat{M}\right)^2\left(2\left(2\hat{q}-\hat{Q}+ \beta\lambda_w\right)\right)^{-1}\right]\right]^n\\
			&= \int D{\xi}\operatorname{ln}\left[\sqrt{\frac{2\pi}{2\hat{q}-\hat{Q}+ \beta\lambda_w}}\exp\left[\frac{\left( {\xi} \sqrt{-\hat{Q}} -\hat{M}\right)^2}{2\left(2\hat{q}-\hat{Q}+ \beta\lambda_w\right) }\right]\right]\\
			& = \frac{1}{2}\operatorname{ln}\frac{2\pi}{ \left(2{\hat{q}}-{\hat{Q}}+ \beta\lambda_w\right)} + \frac{\hat{M}^{2} - \hat{Q}}{2\left(2\hat{q}-\hat{Q}+ \beta\lambda_w\right) }.
		\end{aligned}
	\end{equation}

Before taking the limit of $\Psi_E$, we parameterize the random pre-activation $z_1^a$ and $z_2^a$ as follows:
\begin{equation}
		\begin{aligned}
			&z_1^a = m y_1{M} + \Delta\sqrt{Q}{v_1} +\Delta\sqrt{q-Q}u_1^a,\\ 
			&z_2^a = m y_2{M} + \Delta\sqrt{Q}{v_2} + \Delta\sqrt{q-Q}u_2^a, 
		\end{aligned}
	\end{equation}
	where $v_1,v_2,u_1^a,u_2^a$ are standard Gaussian random variables. The above parameterization obeys the statistics of $(z_1^a,z_2^a)$ [see Eq.~\eqref{covz}]. Therefore, the energy term can be recast into the following form:
	\begin{equation}
		\begin{aligned}
			\Psi_{E} &=\alpha \operatorname{ln} \sum_{\{y_1,y_2\}}\mathrm{P}(y_1,y_2)\int Dv_1Dv_2\int \prod^n_{a}Du_1^aDu_2^a\\
			&\quad\quad\quad\quad\prod^{n}_{a}\operatorname{exp}\left[-\beta\left(\frac{\sigma}{2}\left(D^{a}\right)^2+\frac{1-\sigma}{2} \varphi\left[d_F - \left(D^{a}\right)^2\right] \right)\right],
		\end{aligned}
	\end{equation}
where $D^{a} = \sqrt{\left[\phi\left(z_1^a\right)-\phi\left(z_2^a\right)\right]^2}$. After taking the limit $\lim_{n\to0}\frac{\Psi_E(n)}{n}$, we get
\begin{equation}
		\begin{aligned}
			\Psi_{E} \left({Q}, {q}, {M} \right) = \alpha\sum_{\{{y_1},{y_2}\}}\mathrm{P}(y_1, y_2)\int Dv_1Dv_2 \ln\int Du_1Du_2 \mathrm{exp}\left[-\beta \mathcal{H}\left({Q}, {q}, {M}\right)\right],
				\end{aligned}
	\end{equation}
where the effective single variable Hamiltonian reads
		\begin{equation} \label{eq:H_rs}
			  \mathcal{H}\left({Q}, {q}, {M}\right)= \frac{\sigma}{2}D^2+\frac{1-\sigma}{2} \varphi\left(d_F - D^2\right),
	\end{equation}
	where $\sigma=\frac{1+y_1y_2}{2}$, $D^2=[\phi(z_1)-\phi(z_2)]^2$, $z_x=my_xM+\Delta\sqrt{Q}v_x+\Delta\sqrt{q-Q}u_x$ ($x=1,2$).

Collecting all three contributions to the replica symmetric free energy, we get 
\begin{equation}
-\beta f= \mathop{\mathrm{Extr}}\limits_{M,q, Q,\hat{M},\hat{q},\hat{Q}} {\hat{M}}{M} +{\hat{{q}}}{q} -
			\frac{{\hat{{Q}}}{Q}}{2} 
   +\Psi_{S}\left({\hat{Q}}, {\hat{q}}, {\hat{M}}\right) + \Psi_{E} \left({Q}, {q}, {M} \right),
\end{equation}
which has been shown in the main text.

The self-consistent equations the order parameters must obey are derived by taking the derivative $\frac{\partial[-\beta f]}{\partial\mathcal{O}}$ to be zero, where $\mathcal{O}$ represents all order parameters and their conjugate counterparts. We skip the technical details here, but we remind the readers that to simplify the results, one has to use an integral identity $\int DzzF(z)=\int Dz F'(z)$ (also named the Stein Lemma) which can be proved using integral by parts, and $F'(z)$ is an any differentiable function. The resultant equations are called saddle-point equations (SDEs), which is summarized as follows.
\begin{equation}\label{SDE-main}
	\begin{aligned}
		&\hat{q} =\frac{\alpha\beta\Delta^2}{2} \mathop{\mathbb{E}}\limits_{\mathbf{y},\mathbf{v}} \langle\langle \mathcal{H}_{11} + \mathcal{H}_{22}- \beta(\mathcal{H}_{1}^2+\mathcal{H}_{2}^2)\rangle\rangle,\\
		&\hat{Q} =  -\alpha\beta^2\Delta^2 \mathop{\mathbb{E}}\limits_{\mathbf{y},\mathbf{v}} \left[\langle\langle\mathcal{H}_{1} \rangle\rangle^2 + \langle\langle\mathcal{H}_{2} \rangle\rangle^2 \right],\\
		&\hat{M} = \alpha\beta m \mathop{\mathbb{E}}\limits_{\mathbf{y},\mathbf{v}}\left[ y_1\langle\langle \mathcal{H}_1\rangle\rangle + y_2\langle\langle\mathcal{H}_2 \rangle\rangle\right],\\
		& q = \left(2{\hat{q}}-{\hat{Q}+ \beta\lambda_w}\right)^{-1} + \frac{\hat{{M}}^2 - \hat{{Q}}}{\left({2\hat{q}}-{\hat{Q}+ \beta\lambda_w}\right)^2 },\\
		& Q = \frac{\hat{{M}}^2 - \hat{{Q}}}{\left({2\hat{q}}-{\hat{Q}+ \beta\lambda_w}\right)^2 },\\
		&{M}  = -\frac{\hat{{M}}}{\left({2\hat{q}}-{\hat{Q}+ \beta\lambda_w}\right) },
	\end{aligned}
\end{equation}
where $\mathbf{y}=(y_1,y_2)$, $\mathbf{v}=(v_1,v_2)$, $\mathbf{u}=(u_1,u_2)$, $\mathcal{H}_{i}=\frac{\partial\mathcal{H}}{\partial z_i}$, and $\mathcal{H}_{11}$ and $\mathcal{H}_{22}$ are corresponding second derivatives; $\langle\langle\bullet\rangle\rangle$ denotes the thermal average under the measure of effective Hamiltonian
expressed as $\langle\langle\bullet\rangle\rangle=\frac{\int D\mathbf{u} e^{-\beta\mathcal{H}}\bullet}{\int D\mathbf{u}e^{-\beta\mathcal{H}}}$. The Hamiltonian derivatives are specified by $\mathcal{H}_{1} = \frac{\partial \mathcal{H}}{\partial z_1}$, and $\mathcal{H}_{2} = \frac{\partial \mathcal{H}}{\partial z_2}$, $\mathcal{H}_{11} = \frac{\partial^2 \mathcal{H}}{\partial z_1\partial z_1}$ and $\mathcal{H}_{22} = \frac{\partial^2 \mathcal{H}}{\partial z_2\partial z_2}$. Their explicit expressions are given below:
\begin{equation}
	\begin{aligned}\label{eqn_dH}
		&\mathcal{H}_{1} = \sigma D\phi'_1 - D\left(1-\sigma\right)\varphi^\prime\phi_1^\prime, \\
		&\mathcal{H}_{2} = -\sigma D\phi_2^\prime + \left(1-\sigma\right)\varphi^\prime D\phi_2^\prime, \\
		&\mathcal{H}_{11} = \sigma\left[(\phi_1^\prime)^2+ D\phi_1^{\prime\prime}\right] - \left(1-\sigma\right)\left\{-2D^2\varphi^{\prime\prime}(\phi'_1)^2 + \varphi^{\prime}\left[(\phi_1^\prime)^2+ D\phi_1^{\prime\prime}\right]\right\}, \\
		&\mathcal{H}_{22} = \sigma\left[(\phi_2^\prime)^2 -D\phi_2^{\prime\prime}\right]+ \left(1-\sigma\right)\left\{2\varphi^{\prime\prime}D^2(\phi^\prime_2)^2 - \varphi^{\prime}\left[(\phi_2^\prime)^2 -D\phi_2^{\prime\prime}\right]\right\},\\
	\end{aligned}
\end{equation}
where $\phi_x\equiv\phi(z_x)$ ($x=1,2$), $D=\phi_1-\phi_2$, $\varphi^\prime\left(x\right) = \frac{1}{2}\left(1 + \frac{x}{\sqrt{x^2 + a}}\right)$, $\varphi^{\prime\prime}\left(x\right) = \frac{a}{2\left(x^2+a\right)^{\frac{3}{2}}}$, and the argument of $\varphi'$ (or $\varphi'')$ is $d_F-D^2$.

\subsection{Generalization contrastive loss}
The order parameters of $M,q$ determine the following generalization contrastive loss: 
\begin{equation}
	\begin{aligned}			
 \epsilon_g=\mathbb{E}_{\mathcal{D}}\left[\langle\mathcal{L}^\mu\left(\mathbf{w^*}\right)\rangle\right],		
	\end{aligned}
\end{equation}
where $\langle\cdot\rangle$ means the average over the Boltzmann measure of weights, $\mathcal{L}^\mu= \frac{1}{2}\left[\sigma^\mu D_\mu^2+(1-\sigma^\mu) \varphi\left(d_F - D_\mu ^2\right) \right]$, and $\mathbf{w^*} \in \mathbb{R}^N$ is the optimal weight configuration following the Boltzmann measure. We can then use the fact that given the label $\mathbf{y}$, the pre-activations of $z_1$ and $z_2$ follow a joint Gaussian distribution, and then the average over $\mathbf{x}$ can be integrated out, leading to the following result:
\begin{equation}
		\epsilon_g =\mathbb{E}_{y_1, y_2}\int Du_1\int Du_2 \left[\frac{\sigma}{2} D^2\left(\phi_1, \phi_2 \right)+\frac{1-\sigma}{2} \varphi\left[d_F - D^2\left(\phi_1, \phi_2 \right)\right]\right],
\end{equation}
where $D^2(\phi_1,\phi_2)=[\phi_1(z_1)-\phi_2(z_2)]^2$, and $z_x\equiv my_xM+\Delta\sqrt{q}u_x$ ($x=1,2$). 

\subsection{Distance of fermion pair}
As in the derivation of generalization contrastive loss, the distance of fermion pair can also be estimated in an analytic form:
\begin{equation}
        D^2_{F} =\int Du_1Du_2 \left[\phi\left( mM + \Delta{\sqrt{q}u_1}\right) - \phi\left( -mM + \Delta{\sqrt{q}u_2}\right)\right]^2,
\end{equation}
where we have assumed a fermion pair $(y_1,y_2)=(+1,-1)$ is considered, and the result is the same in the case of $(y_1,y_2)=(-1,+1)$.

\subsection{Generalization accuracy}
\begin{figure}
    \centering
\includegraphics[scale=0.9]{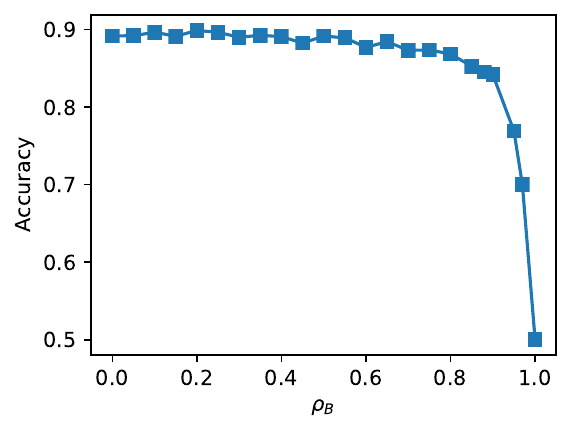}
        \caption{Generalization accuracy vs. the fraction of Boson pair. Other parameters are the same as in Fig. 2 in the main text. } 
        \label{figrho}
    \end{figure}

We use a sign function readout to measure how good the contrastive representation learning is. If the hidden representation (in our toy model this
is one dimension latent space) is well separated, then the error rate for a test data point can be defined as follows,
\begin{equation}
    \begin{aligned}
        \epsilon_g^C = \left\langle \Theta\left[-yz\left(\mathbf{x},\mathbf{w}\right)\right]\right\rangle_{\mathbf{w},\mathcal{D}},
    \end{aligned}
\end{equation}
where $\langle\cdot \rangle$ means the average over the test data and the optimal weight configuration following the Boltzmann measure. Note that the pre-activation $z$ is a Gaussian random 
variable $\mathcal{N}(z;Mmy,q\Delta^2)$. We also assume that the two clusters corresponding to two classes are correctly separated with corresponding labels ($y\to-y$ otherwise).
We have then the following derivation:
\begin{equation}\label{stdacc}
    \begin{aligned}
       \epsilon_g^C &=\mathbb{E}_y\int Du\Theta(-y(Mmy+\Delta\sqrt{q}u))\\
       &=\int Du\Theta(-Mm+\Delta\sqrt{q}u)\\
       &=\frac{1}{2}\left[1-\mathrm{erf}\left(\frac{Mm}{\Delta\sqrt{2q}}\right)\right],
    \end{aligned}
\end{equation}
where $m=1$ in our Gaussian mixture data setting, and $C$ indicates classification. We can thus get the generalization accuracy as $1-\epsilon^C_g$, which is plotted as a function of $\rho$ in Fig.~\ref{figrho}. The accuracy deteriorates sharply when the fraction of Boson pairs in the training dataset goes above a threshold.

\subsection{Theory of adversarial attack on FBM}
Because the FGSM attack leads to an order parameter $\Lambda=\sum_i{|w_i|}/N\leq\sqrt{q}$, we would not analyze the FGSM attack using the replica theory of FBM. Instead, we analyze the $\ell_2$ norm attack defined as follows:
\begin{equation}
    \begin{aligned}
        \delta \mathbf{x} = {\epsilon} \frac{\nabla_{\mathbf{x}}\mathcal{L}}{\sqrt{\sum_i\left(\nabla_{{x_i}}\mathcal{L}\right)^2}},
    \end{aligned}
\end{equation}
where $ {\epsilon}$ is the attack strength. Note that FGSM is an $\ell_\infty$ norm attack~\cite{Good-2015,Madry-2018}. We conjecture that the theory of $\ell_2$ norm attack would qualitatively predict behavior of a broad range of $\ell_p$ norm bounded attacks including FGSM. The attack is calculated from a mean-squared loss specified as follows:
\begin{equation}
    \begin{aligned}
        \mathcal{L} = \frac{1}{2}\left[y - \phi\left(\sum_iw_ix_i\right)\right]^2.
    \end{aligned}
\end{equation}
Therefore, one can compute the increment of the preactivation caused by the attack, i.e., $\mathbf{w}^\t\delta\mathbf{x}$, which reads
\begin{equation}
    \begin{aligned}
\mathbf{w}^\t\delta\mathbf{x} &= \epsilon\mathbf{w}^\t\frac{-\left[y-\phi\left(z\right)\right]\phi^{\prime}\left(z\right)\mathbf{w}}{\sqrt{[-\left[y-\phi\left(z\right)\right]\phi^{\prime}\left(z\right)]^2\mathbf{w}^\t\mathbf{w}}}\\
&=\epsilon \frac{\mathcal{K}_y\left(z\right)q}{\sqrt{\mathcal{K}^2_y\left(z\right)q}}\\
&=\epsilon\sqrt{q}\mathrm{sgn}\left[\mathcal{K}_y\left(z\right)\right],
    \end{aligned}
\end{equation}
where we have rescaled the attack strength to make the increment of the order one, keeping the same order with $z=\mathbf{w}^\t\mathbf{x}$, and $\mathcal{K}_y\left(z\right) = -\left[y-\phi\left(z\right)\right]\phi^{\prime}\left(z\right)$.  $q \equiv \frac{\mathbf{w}^\t\mathbf{w}}{N}$, which can be treated as the order parameter in our replica theory. The averaged adversarial accuracy as a function of attack strength is thus given by
\begin{equation}
    \begin{aligned}
    {\rm ACC}_{adv} &= 1 - \left\langle \Theta\left[-y\left(z + \mathbf{w}^\t\delta\mathbf{x}\right)\right]\right\rangle_{\mathbf{w},x,y}\\
    &= 1 - \left\langle \Theta\left(-yz -y \epsilon\sqrt{q} \mathrm{sgn}\left[\mathcal{K}_y\left(z\right)\right]\right)\right\rangle_{z,y}\\
    &=1-\frac{1}{2}\int Du\Theta\left(-mM -\Delta\sqrt{q} u - \epsilon\sqrt{q}\mathrm{sgn}\left[\mathcal{K}_1\left(mM +\Delta\sqrt{q} u\right)\right]\right)\\
        &\quad-\frac{1}{2}\int Du\Theta\left(-mM + \Delta\sqrt{q} u + \epsilon\sqrt{q}\mathrm{sgn}\left[\mathcal{K}_{-1}\left(-mM + \Delta\sqrt{q} u\right)\right]\right)\\
        &=\frac{1}{2}\int Du\Theta\left(mM +\Delta\sqrt{q} u +\epsilon\sqrt{q}\mathrm{sgn}\left[\mathcal{K}_1\left(mM +\Delta\sqrt{q} u\right)\right]\right)\\
        &\quad+\frac{1}{2}\int Du\Theta\left(mM -\Delta\sqrt{q} u - \epsilon\sqrt{q}\mathrm{sgn}\left[\mathcal{K}_{-1}\left(-mM + \Delta\sqrt{q} u\right)\right]\right)\\
         &=\frac{1}{2}\int Du\Theta\left(mM +\Delta\sqrt{q} u +\epsilon\sqrt{q}\mathrm{sgn}\left[\mathcal{K}_1\left(mM +\Delta\sqrt{q} u\right)\right]\right)\\
        &\quad+\frac{1}{2}\int Du\Theta\left(mM +\Delta\sqrt{q} u - \epsilon\sqrt{q}\mathrm{sgn}\left[\mathcal{K}_{-1}\left(-mM -\Delta\sqrt{q} u\right)\right]\right)\\
        &=\int Du\Theta\left(mM +\Delta\sqrt{q} u +\epsilon\sqrt{q}\mathrm{sgn}\left[\mathcal{K}_1\left(mM +\Delta\sqrt{q} u\right)\right]\right),
      \end{aligned}
\end{equation}
where we have used $z \sim \mathcal{N}\left(mMy, \Delta^2 q\right)$, $\mathrm{P}\left(y=\pm 1\right)=\frac{1}{2}$, $\Theta(x)=1-\Theta(-x)$, $\mathcal{K}_{-1}(-z)=-\mathcal{K}_1(z)$, and $u$ is symmetric for the standard Gaussian distribution. Note that $\phi(\cdot)=\tanh(\cdot)$ in this paper for the toy model. We remark that the averaged adversarial accuracy in the absence of attack reduces to the standard accuracy derived in Eq.~\eqref{stdacc}, i.e., $\int Du\Theta(mM+\sqrt{q}\Delta u)$.

\subsection{Zero temperature}
The zero temperature limit would make the Boltzmann measure focus on the ground state of lowest energy. In this limit, the order parameters including their conjugate counterparts will diverge or vanish with $\beta$ as $\beta\to\infty$. Therefore, the (conjugated) order parameters need to be properly scaled with $\beta$.  We have then to check the finite temperature free energy first and determine the following scaling behavior of order parameters:
\begin{equation}
	\begin{aligned}
		&\left(q-Q\right)\to\frac{\chi} {\beta}, \\
		&\left(2\hat{q} - \hat{Q}\right)\to \beta\hat{\chi}, \\
		&\hat{M} \to \beta\hat{M}, \\
		&\hat{Q} \to \beta^2\hat{Q}. \\
	\end{aligned}
\end{equation}
Therefore, we estimate the limit $\lim_{\beta\to\infty}\frac{\Psi_0+\Psi_S+\Psi_E}{\beta}$. The first limit of $\Psi_0$ is obtained as
\begin{equation}
\Psi_0^\infty=\frac{1}{2}\left(q\hat{\chi}-\chi\hat{Q}\right)+M\hat{M},
\end{equation}
and the second limit of $\Psi_S$ is obtained as
	\begin{equation}
	\Psi_{S}^\infty =  \frac{\hat{{M}}^{2} - \hat{{Q}}}{2\left(\hat{\chi}+ \lambda_w\right) },
\end{equation}
and the final limit of $\Psi_E$ reads
\begin{equation}\label{minF}
\begin{aligned}
	\Psi_{E}^\infty &= -\alpha\mathbb{E}_{\mathbf{y}}\int D\mathbf{v}F_{\mathbf{y}}(\mathbf{v}),\\
	F_{\mathbf{y}}(\mathbf{v})&=\min_{u_1,u_2}\left[\frac{u_1^2}{2}+\frac{u_2^2}{2}+\mathcal{H}(y_1,y_2,z_1,z_2)\right],\\
	z_x&=my_xM+\sqrt{\Delta^2q}v_x+\sqrt{\Delta^2\chi}u_x,\\
	\end{aligned}
\end{equation}
 where $x=1,2$. Collecting all three contributions in the zero temperature limit, we obtain the ground state free energy as follows:
 \begin{equation}
-f(T=0)=\Psi_0^\infty+\Psi_S^\infty+\Psi_E^\infty.
\end{equation}

Now, the introduced new set of order parameters, $\{q,\hat{Q},M,\hat{M},\chi,\hat{\chi}\}$ must obey the saddle-point equations which can be derived in an analogous way to the finite temperature version shown in this section. The SDEs now become
\begin{equation}  \label{eq:zeroT_rs}
	\begin{aligned}
		&M = -\frac{\hat{M}}{\hat{\chi} + \lambda_w},\\
		&q = \frac{\hat{M}^2-\hat{Q}}{\left(\hat{\chi} + \lambda_w\right)^2}, \\
		&\chi =- \frac{1}{\hat{\chi}+\lambda_w},\\
		&\hat{M} =\alpha\mathbb{E}_{\mathbf{y}}\int D\mathbf{v}\left(\mathcal{H}_1^*my_1+\mathcal{H}_2^*my_2\right),\\
            &\hat{Q} =-\frac{\alpha\Delta^2}{\sqrt{\Delta^2\chi}}\mathbb{E}_{\mathbf{y}}\int D\mathbf{v}\left(\mathcal{H}_1^*u_1^*+\mathcal{H}_2^*u_2^*\right),\\
            &\hat{\chi} = \alpha\Delta^2 \mathbb{E}_{\mathbf{y}}\int D\mathbf{v}\left(\mathcal{H}_{11}^*+\mathcal{H}_{22}^*\right),    
	\end{aligned}
\end{equation}
where the superscript $*$ means that the optimal values of $u_1$ and $u_2$ obtained by solving the min problem in Eq.~\eqref{minF} are used to estimate the effective Hamiltonian, and to arrive at the last equation, we use the Stein Lemma.

\section{Cavity method for FBM}
In this section, we derive the message passing equations applied to single instances of learning, using the cavity method developed in spin glass theory~\cite{Mezard-1987,HH-2022}. These equations are used to verify our theoretical results of replica calculation ($N\to\infty$) on finite-sized systems.
The two cavity probabilities defined in the main text obey the following self-consistent equations under the seminal cavity approximation in disordered system theory~\cite{Mezard-1987,HH-2022}:
\begin{equation}\label{BP-app}
	\begin{aligned}
		&P_{i\to\nu}\left(w_i\right)\propto e^{-\frac{\beta\lambda_w w_i^2}{2}}\prod_{\mu\in\partial i\backslash\nu}\hat{P}_{\mu\to i}\left(w_i\right),\\
		&\hat{P}_{\mu\to i}\left(w_i\right)=\int\prod_{j\in\partial\mu\backslash i}\mathrm{d}w_jP_{j\to\mu}\left(w_j\right)P_{\mu}\left(\mathbf{y}^\mu,\mathbf{z}^\mu\right),
	\end{aligned}
\end{equation}
where $\mathbf{z}^\mu=(z^\mu_1, z^\mu_2)$ are the pre-activations triggered by $\mathbf{x}_1^\mu, \mathbf{x}_2^\mu$, respectively, and $P_{\mu}= e^{-\beta \mathcal{L}_\mu}$ ($\mathcal{L}_\mu$ is the $\mu$-th pair related loss). The explicit expression of $\mathcal{L}^\mu$ is given below.
\begin{equation}
	\begin{aligned}
		&\mathcal{L}^\mu =\frac{\sigma^\mu}{2}\left(D^{\mu}\right)^2+\frac{1-\sigma^\mu}{2} \varphi\left[\lambda_F - \left(D^{\mu} \right)^2\right],\\
		&D^\mu = \sqrt{\left[\phi\left(z_1^{\mu}\right)- \phi\left(z_2^{\mu}\right)\right]^2},\\
		&\sigma^\mu = \frac{1+y^\mu_1y^\mu_2}{2}.
	\end{aligned}
\end{equation}
The prior distribution of weight $\mathrm{P}_0(\mathbf{w})=\prod_i\mathrm{P}_0(w_i)$ is a Gaussian distribution (i.e., the standard $\ell_2$ norm weight decay), as shown below.
\begin{equation}
	\begin{aligned}
		\mathrm{P}_0\left(w_i\right) \propto \mathrm{exp}\left(-\frac{\beta\lambda_w w^2_i}{2}\right),
	\end{aligned}
\end{equation}
where $\lambda_w$ controls the strength of weight decay. 

Because of the continuous nature of weights, we first assume a Gaussian distribution $\mathcal{N}(w_i;m_{i\to\mu},v_{i\to\mu})$ for $P_{i\to\mu}(w_i)$ whose mean
and variance are given below.
\begin{equation}
	\begin{aligned}
		m_{j\rightarrow\mu}& =\int\mathrm{d}w_jw_jP_{j\rightarrow\mu}\left(w_j\right),  \\
		v_{j\rightarrow\mu}& =\int\mathrm{d}w_jw_j^2P_{j\rightarrow\mu}\left(w_j\right)-m_{j\rightarrow\mu}^2. 
	\end{aligned}
\end{equation}
We use this approximation to simplify the conjugate cavity probability $\hat{P}_{\mu\to i}\left(w_i\right)$ as follows.
\begin{equation}\label{phat}
	\begin{aligned}
		\hat{P}_{\mu\to i}\left(w_i\right)&=\int\prod_{j\in\partial\mu\backslash i}\mathrm{d}w_jP_{j\to\mu}\left(w_j\right)P_{\mu}\left(\mathbf{y}^\mu,\mathbf{z}^\mu\right),\\
        &=\sqrt{\frac{4\pi^2}{\mathrm{det}\left(\mathbf{V}_{\mu\to i}\right)}} \int \mathrm{d}z^{\mu}_1 \mathrm{d}z^{\mu}_2 P_{\mu}\left(\mathbf{y}^{\mu},z_1^\mu, z_2^\mu\right) e^{-\frac{1}{2}\left(\mathbf{z}^\mu -w_i \mathbf{x}^\mu_i - \boldsymbol{\omega}_{\mu\rightarrow i}\right)^\t\mathbf{V}^{-1}_{\mu\rightarrow i}\left(\mathbf{z}^\mu -w_i \mathbf{x}^\mu_i - \boldsymbol{\omega}_{\mu\rightarrow i}\right) },
	\end{aligned}
\end{equation}
where we have defined $\mathbf{z}^\mu=(z^\mu_1, z^\mu_2)$, $\mathbf{x}^\mu_i = \left[{x}^{\mu}_{1,i}, {x}^{\mu}_{2,i}\right]^\t$, $\boldsymbol{\omega}_{\mu \rightarrow i} = \sum_{j\in\partial\mu\backslash i} m_{j\rightarrow\mu}\mathbf{x}^\mu_{j}$, and
		$\mathbf{V}_{\mu\rightarrow i} =  \sum_{j\in\partial\mu\backslash i} v_{j\rightarrow\mu}\mathbf{x}^\mu_{j}(\mathbf{x}^\mu_{j})^\t$.

Next, we approximate the integral in Eq.~\eqref{phat} (defined as $I_\mu$) as follows,
\begin{equation}\label{eq:Z_fun}
	\begin{aligned}
		I_\mu &= \int \mathrm{d}z_1^\mu\mathrm{d}z_2^\mu  P_{\mu}\left(\mathbf{y}^{\mu}, z_1^\mu, z_2^\mu\right)  \exp\left[-\frac{1}{2}\left(\mathbf{z}^\mu -w_i \mathbf{x}^\mu_i - \boldsymbol{\omega}_{\mu\rightarrow i}\right)^\t\mathbf{V}^{-1}_{\mu\rightarrow i}\left(\mathbf{z}^\mu -w_i \mathbf{x}^\mu_i - \boldsymbol{\omega}_{\mu\rightarrow i}\right) \right]\\
		&\simeq \int \mathrm{d}z_1^\mu \mathrm{d}z_2^\mu P_{\mu}\left(\mathbf{y}^{\mu}, z_1^\mu, z_2^\mu\right)
		\mathrm{exp}\left[-\frac{1}{2}\left(\mathbf{z}^\mu - \boldsymbol{\omega}_{\mu\rightarrow i}\right)^T\mathbf{V}^{-1}_{\mu\rightarrow i}\left(\mathbf{z}^\mu - \boldsymbol{\omega}_{\mu\rightarrow i}\right) \right]\\
		&\times \left[1 + w_i\left(\mathbf{z}^\mu - \boldsymbol{\omega}_{\mu\rightarrow i}\right)^\t\mathbf{V}^{-1}_{\mu\rightarrow i}\mathbf{x}^\mu_i -\frac{1}{2}\left(w_i\right)^2\left(\mathbf{x}^\mu_i\right)^\t\mathbf{V}^{-1}_{\mu\rightarrow i}\mathbf{x}^\mu_i\right.\\
		 &\left.+\frac{1}{2}w^2_i \left(\mathbf{x}^\mu_i\right)^\t\mathbf{V}^{-1}_{\mu\rightarrow i}\left(\mathbf{z}^\mu - \boldsymbol{\omega}_{\mu\rightarrow i}\right)\left(\mathbf{z}^\mu - \boldsymbol{\omega}_{\mu\rightarrow i}\right)^\t\mathbf{V}^{-1}_{\mu\rightarrow i}\mathbf{x}^\mu_i \right],
	\end{aligned}
\end{equation}
where the last approximation is based on the large $N$ limit. More precisely, in the large $N$ limit, we assume the entries of matrices $\mathbf{A,C}$ are of the order one, but  entries of $\mathbf{B}$ are of the order $\mathcal{O}(1/N)$. Then we have the following mathematical result:
\begin{equation}\label{appro}
\exp\left[-\frac{1}{2}(\mathbf{A}-\mathbf{B})^\t\mathbf{C}(\mathbf{A}-\mathbf{B})\right]\simeq\exp\left[-\frac{1}{2}\mathbf{A}^\t \mathbf{C}\mathbf{A}\right]\left[1+\mathbf{B}^\t\mathbf{C}\mathbf{A}-\frac{1}{2}\mathbf{B}^\t\mathbf{C}\mathbf{B}+\frac{1}{2}\mathbf{B}^\t\mathbf{C}\mathbf{A}\mathbf{A}^\t\mathbf{C}^\t\mathbf{B}\right],
\end{equation}
where the order higher than $1/N^2$ has been dropped off in the above expansion, and $\mathbf{C}$ is a symmetric matrix. Equation~\eqref{appro} is used to derive the last equality in Eq.~\eqref{eq:Z_fun}.

 From Eq.~\eqref{eq:Z_fun}, we extract a probability measure $\tilde{P}_\mu=\frac{1}{\tilde{Z}_\mu}e^{-\frac{1}{2}(\mathbf{z}^\mu-\boldsymbol{\omega}_{\mu\to i})^\t \mathbf{V}^{-1}_{\mu\to i}(\mathbf{z}^\mu-\boldsymbol{\omega}_{\mu\to i})}P_\mu$, from which we derive $\mathbf{f}_{\mu}\equiv\frac{\partial\ln\tilde{Z}_\mu}{\partial\boldsymbol{\omega}_{\mu\to i}}$. More precisely, $\mathbf{f}_\mu$ and its derivative with respect to $\boldsymbol{\omega}_{\mu\to i}$ can be derived below.
 \begin{equation}\label{eq:f_fun}
	\begin{aligned}
		\mathbf{f}_\mu &=\frac{\int d\mathbf{z}^\mu \mathbf{V}^{-1}_{\mu\rightarrow i}\left(\mathbf{z}^\mu - \boldsymbol{\omega}_{\mu\rightarrow i}\right)\mathrm{exp}\left[-\frac{1}{2}\left(\mathbf{z}^\mu - \boldsymbol{\omega}_{\mu\rightarrow i}\right)^\t\mathbf{V}^{-1}_{\mu\rightarrow i}\left(\mathbf{z}^\mu - \boldsymbol{\omega}_{\mu\rightarrow i}\right) \right]P_{\mu}\left(\mathbf{y}^\mu,\mathbf{z}^\mu\right)}{\int d\mathbf{z}^\mu \mathrm{exp}\left[-\frac{1}{2}\left(\mathbf{z}^\mu - \boldsymbol{\omega}_{\mu\rightarrow i}\right)^\t\mathbf{V}^{-1}_{\mu\rightarrow i}\left(\mathbf{z}^\mu - \boldsymbol{\omega}_{\mu\rightarrow i}\right) \right]P_{\mu}\left(\mathbf{y}^\mu,\mathbf{z}^\mu\right)},\\
		&\frac{\partial \mathbf{f}_\mu}{\partial_{\boldsymbol{\omega}_{\mu\to i}}} = -\mathbf{V}^{-1}_{\mu\rightarrow i}-\mathbf{f}_\mu (\mathbf{f}_\mu)^\t\\
		&+ \mathbf{V}^{-1}_{\mu\rightarrow i}\frac{\int d\mathbf{z}^\mu\left(\mathbf{z}^\mu - \boldsymbol{\omega}_{\mu\rightarrow i}\right)\left(\mathbf{z}^\mu - \boldsymbol{\omega}_{\mu\rightarrow i}\right)^\t \mathrm{exp}\left[-\frac{1}{2}\left(\mathbf{z}^\mu - \boldsymbol{\omega}_{\mu\rightarrow i}\right)^\t\mathbf{V}^{-1}_{\mu\rightarrow i}\left(\mathbf{z}^\mu - \boldsymbol{\omega}_{\mu\rightarrow i}\right) \right]P_{\mu}(\mathbf{y}^\mu,\mathbf{z}^\mu)}{\int d\mathbf{z}^\mu \mathrm{exp}\left[-\frac{1}{2}\left(\mathbf{z}^\mu - \boldsymbol{\omega}_{\mu\rightarrow i}\right)^\t\left(\mathbf{z}^\mu - \boldsymbol{\omega}_{\mu\rightarrow i}\right) \right]P_{\mu}(\mathbf{y}^\mu,\mathbf{z}^\mu)}\mathbf{V}^{-1}_{\mu\rightarrow i}.\\
		\end{aligned}
	\end{equation}
Based on the above two auxiliary quantities, the conjugate cavity probability $\hat{P}_{\mu\to i}(w_i)$ can be recast as
\begin{equation}\label{hatP}
	\begin{aligned}
		\hat{P}_{\mu\to i}\left(w_i\right)&\propto 1 + w_i \left(\mathbf{x}^\mu_i\right)^\t	\mathbf{f}_\mu + \frac{1}{2}w_i^2\left(\mathbf{x}^\mu_i\right)^\t\left(\partial_{\boldsymbol{\omega}_{\mu\to i}}\mathbf{f}_\mu+\mathbf{f}_\mu\mathbf{f}_\mu^\t \right)\mathbf{x}^\mu_i\\
		&=1+w_iB_{\mu\to i}-\frac{1}{2}w_i^2A_{\mu\to i}+\frac{1}{2}w_i^2B^2_{\mu\to i}\\
		&\simeq\exp\left[w_iB_{\mu\to i}-\frac{1}{2}w_i^2A_{\mu\to i}\right],
	\end{aligned}
\end{equation}
where $B_{\mu \to i} = (\mathbf{x}_i^\mu)^\t\mathbf{f}_\mu$, and 
		$A_{\mu\to i}=-(\mathbf{x}_i^\mu)^\t\frac{\partial\mathbf{f}_\mu}{\partial\boldsymbol{\omega}_{\mu\to i}}\mathbf{x}_i^\mu$.
Inserting Eq.~\eqref{hatP} into the first equation of the message passing equations, and noting that $P_{i\to\mu}\sim\mathcal{N}(m_{i\to\mu},v_{i\to\mu})$, one can derive self-consistently the following closed equations for the cavity mean and variance.
\begin{equation}\label{BP-final}
	\begin{aligned}
		&m_{i \to \mu}= \frac{\sum_{\nu\neq\mu}B_{\nu\to i}}{\beta\lambda_w+\sum_{\nu\neq\mu}A_{\nu\to i}}, \\	
		&v_{i \to \mu} =\frac{1}{\beta\lambda_w+\sum_{\nu\neq\mu}A_{\nu\to i}}.
		\end{aligned}
		\end{equation}
A full version of the mean and variance is derived by restoring the $\mu$-th data pair as follows,
\begin{equation}\label{BP-full}
	\begin{aligned}
		&m_{i}= \frac{\sum_{\nu\in\partial i}B_{\nu\to i}}{\beta\lambda_w+\sum_{\nu\in\partial i}A_{\nu\to i}}, \\	
		&v_{i} =\frac{1}{\beta\lambda_w+\sum_{\nu\in\partial i}A_{\nu\to i}}.
		\end{aligned}
		\end{equation}
Finally, $P_i(w_i)=\mathcal{N}(w_i;m_i,v_i)$ from which we can estimate the generalization contrastive loss (see Fig. 2 in the main text).


\end{document}